%% file: main.tex
\documentclass{article}

\PassOptionsToPackage{numbers,compress}{natbib}
\usepackage[preprint]{neurips_2026}

\usepackage[utf8]{inputenc}
\usepackage[T1]{fontenc}
\usepackage{hyperref}
\usepackage{url}
\usepackage{graphicx}
\usepackage{booktabs}
\usepackage{bbm}
\usepackage{colortbl}
\usepackage{caption}
\usepackage{amsmath}
\usepackage{amsfonts}
\usepackage{amssymb}
\usepackage{nicefrac}
\usepackage{microtype}
\usepackage{xcolor}
\usepackage{listings}
\usepackage{wrapfig}
\usepackage{multirow}
\usepackage{tcolorbox}
\usepackage{fontawesome5}
\tcbuselibrary{listings,skins,breakable}
\usepackage{pifont} 

\definecolor{promptbg}{HTML}{F7F7F4}
\definecolor{promptframe}{HTML}{B8B8B0}
\definecolor{promptkey}{HTML}{1F6FB2}
\definecolor{promptstr}{HTML}{8B5A2B}
\definecolor{promptcom}{HTML}{6E7781}
\definecolor{promptplaceholder}{HTML}{C0392B}

\usepackage{tikz}

\newsavebox{\eyeon}
\savebox{\eyeon}{%
  \begin{tikzpicture}[baseline=-0.6ex, line cap=round, line join=round]
    \draw[line width=0.5pt] (0,0) ellipse (3.2pt and 1.9pt);
    \fill (0,0) circle (1pt);
  \end{tikzpicture}%
}

\newsavebox{\eyeoff}
\savebox{\eyeoff}{%
  \begin{tikzpicture}[baseline=-0.6ex, line cap=round, line join=round]
    \draw[line width=0.5pt] (0,0) ellipse (3.2pt and 1.9pt);
    \fill (0,0) circle (1pt);
    \draw[line width=0.7pt] (-3.8pt,-2.4pt) -- (3.8pt,2.4pt);
  \end{tikzpicture}%
}

\newcommand{\rgon}{\usebox{\eyeon}}
\newcommand{\rgoff}{\usebox{\eyeoff}}

\input{figs}

\input{tabs}

\title{Motion-o: Trajectory-Grounded Video Reasoning}

\author{%
  Bishoy Galoaa* \\
  Northeastern University \\
  \texttt{galoaa.b@northeastern.edu}
  \And
  Shayda Moezzi* \\
  Northeastern University \\
  \texttt{moezzi.s@northeastern.edu}
  \And
  Xiangyu Bai \\
  Northeastern University \\
  \texttt{bai.xiang@northeastern.edu}
  \And
  Sarah Ostadabbas \\
  Northeastern University \\
  \texttt{s.ostadabbas@northeastern.edu}
}

\begin{document}

\maketitle

\vspace{-0.5em}
\begin{center}
    \includegraphics[width=0.78\linewidth]{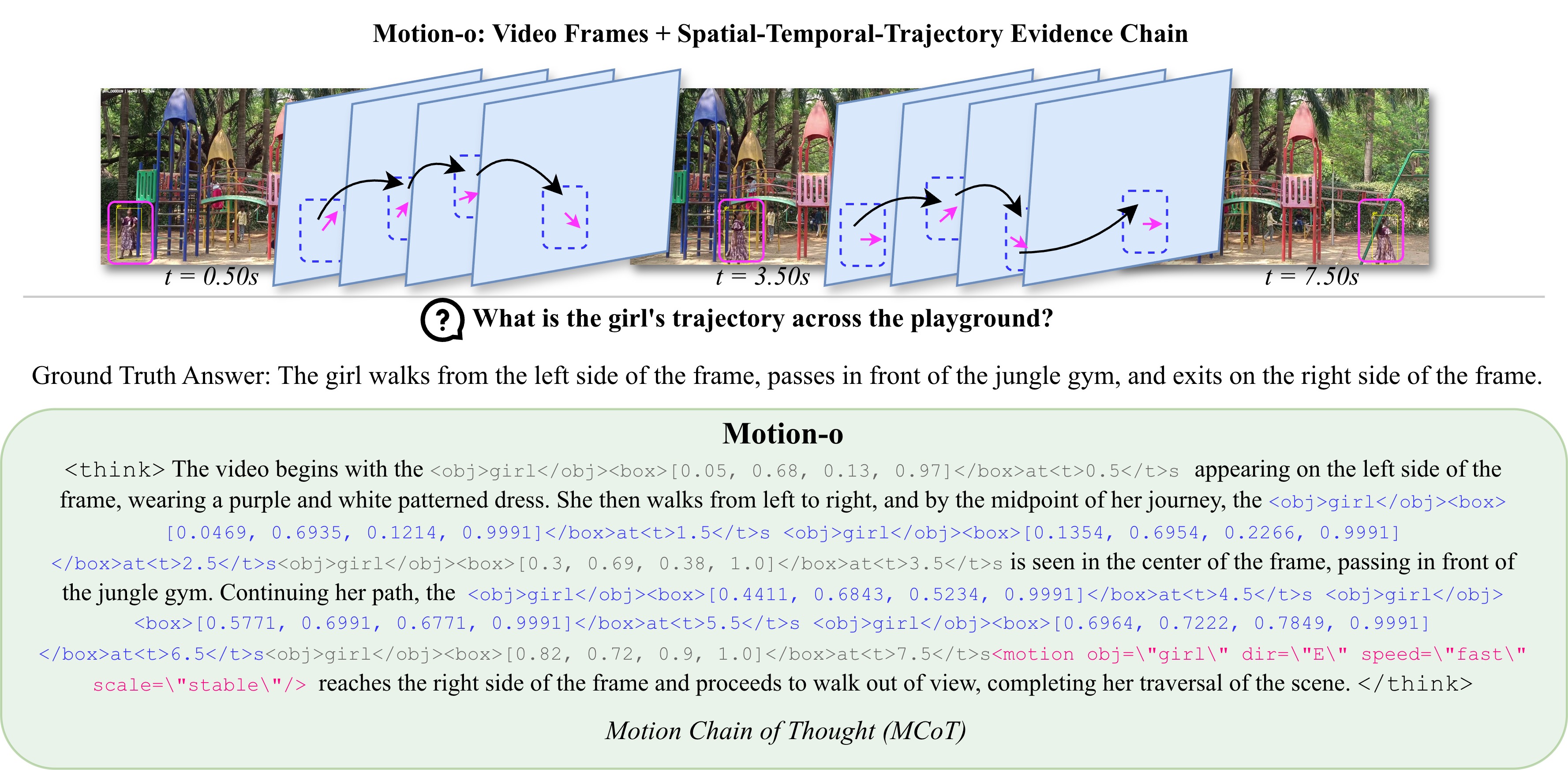}
    \captionsetup{hypcap=false, width=0.82\linewidth}
    \captionof{figure}{Although recent video models can produce fluent reasoning traces, their reasoning is typically ungrounded. Motion-o introduces a \textbf{Spatial--Temporal--Trajectory (STT) evidence chain} that explicitly embeds model reasoning with a trajectory-faithful trace.}
    \label{fig:teaser}
\end{center}
\vspace{1.0em}

\begin{abstract}
    Recent video reasoning models increasingly produce spatio-temporal evidence chains that localize objects at specific timestamps. While these traces improve interpretability by grounding \emph{where} and \emph{when} evidence appears, they often leave the motion connecting observations, the \textit{how}, implicit. This makes dynamic and trajectory-dependent claims difficult to supervise, verify, or penalize when unsupported by the video. We formalize this missing component as Spatial-Temporal-Trajectory (STT) reasoning and introduce \textbf{Motion-o}, a motion-centric extension to vision-language models (VLMs) that makes trajectories explicit and verifiable. Motion-o augments evidence chains with Motion Chain of Thought (MCoT), a structured pathway that represents object motion through a discrete \texttt{<motion/>} tag summarizing direction, speed, and scale change. To supervise MCoT, we densify sparse spatio-temporal annotations into object tracks and derive motion descriptors from centroid displacement and box-area change. We then train with complementary rewards for trajectory consistency and visual grounding, including a perturbation-based signal that penalizes motion descriptions that remain unchanged when temporal evidence is removed. Across multiple video understanding benchmarks, Motion-o consistently improves trajectory-faithful reasoning without architectural modifications. These results suggest that an explicit motion interface can complement existing VLM pipelines by converting implicit dynamics into verifiable evidence. Code is available at~\href{https://github.com/ostadabbas/Motion-o}{\faGithub\ \texttt{ostadabbas/Motion-o}}.
\end{abstract}

\section{Introduction}
\label{sec:intro}
\comparison
Humans effortlessly perceive motion. We recognize when a person walks leftward, when a ball accelerates, or when a vehicle follows a smooth arc. This capacity for motion understanding is so fundamental that it often goes unnoticed, yet it underpins our ability to predict, explain, and reason about the dynamic visual world. However, many current vision-language model (VLM) reasoning pipelines still represent evidence primarily as localized snapshots, and lack an explicit mechanism to reason about how objects move between observations. As a result, models frequently rely on textual interpolation to ``fill in'' dynamics, producing motion statements that are plausible but ungrounded and difficult to verify.

 Recent work on grounded video reasoning has made progress toward structured visual evidence ~\cite{feng2025video,wang2025videorft,meng2025open}. As illustrated in the left example in Figure~\ref{fig:comparison}, such traces improve interpretability by grounding objects in space and time. However, they remain snapshot-based: an \texttt{<obj>} and \texttt{<box>} at a few timestamps specify where an entity appears, but not the trajectory that connects those observations. The missing link is not motion as a low-level visual signal, which has been studied extensively in tracking, robotics, and video generation. Rather, the missing link is motion as an explicit and rewardable component of the model's reasoning trace. Without this component, a model may answer a trajectory-dependent question correctly or incorrectly while leaving no structured evidence for whether its motion claim follows from the visual dynamics~\cite{trove,escalatorproblem,liu2024tempcompass}.

We formalize this missing capability as \textbf{Spatial--Temporal--Trajectory (STT) reasoning}: the joint ability to reason about \emph{where} objects are, \emph{when} they appear, and \emph{how} they move between observations. Existing evidence-based video reasoning frameworks largely address the spatial and temporal dimensions, but leave trajectory information implicit. This is limiting because video understanding requires explicit motion understanding: whether an object moved leftward, remained stationary, approached the camera, changed scale, or followed a curved path. Such descriptors provide the semantic link between discrete observations and dynamic interpretation, supporting motion-sensitive question answering, anomaly detection, trajectory prediction, and embodied reasoning.

We introduce \textbf{Motion-o}, a lightweight extension for making object motion explicit and verifiable in grounded video reasoning. Motion-o augments spatio-temporal evidence chains with \textbf{Motion Chain-of-Thought (MCoT)}, a structured reasoning step that summarizes the motion connecting grounded observations; figure~\ref{fig:teaser} provides an overview of Motion-o and MCoT. After grounding an object at multiple timestamps, the model emits a compact self-closing \texttt{<motion/>} tag with discrete, interpretable attributes for direction, speed, and scale change; for example, \texttt{<motion obj=``person'' dir=``W'' speed=``slow'' scale=``stable''/>}. This converts an implicit trajectory into a structured intermediate statement that can be parsed, supervised, rewarded, and inspected. 


The central question of this work is not whether motion can be represented in principle. Prior work has explored visual traces for robotics, trajectory-based controls for generation, and spatio-temporal grounding for localization. Motion-o targets a different interface: open-ended video reasoning, where the model must connect localized observations into a trajectory-faithful explanation and final answer. The value of MCoT is therefore not that direction, speed, or scale are novel primitives in isolation, but that these primitives become structured reasoning variables inside the VLM output.


To train Motion-o, we augment existing spatio-temporal annotations with denser object tracks and derive discrete motion descriptors from centroid displacement and box-area change. These descriptors teach the MCoT format during supervised fine-tuning, while trajectory and visual-grounding rewards encourage predicted motion tags to both match the observed track and depend on temporal visual evidence. Motion-o is intentionally simple and architecture-agnostic: it exposes object motion as a structured intermediate variable that can be supervised and verified within existing VLM pipelines.

Our contributions are as follows: (1) we formalize \textbf{Spatial--Temporal--Trajectory (STT) reasoning} as an extension of grounded video reasoning from localized snapshots to object motion between observations; (2) we introduce \textbf{Motion Chain-of-Thought (MCoT)}, a structured \texttt{<motion/>} reasoning operator that encodes motion as parseable intermediate evidence; (3) we propose \textbf{Motion-o}, an architecture-agnostic training framework that teaches and rewards explicit motion reasoning through supervised fine-tuning and trajectory-aware reinforcement learning; and (4) we construct a trajectory-grounding augmentation of existing spatio-temporal annotations, deriving dense object tracks and motion descriptors for supervising trajectory-faithful video reasoning.

\section{Related Work}
\label{sec:related}

\textbf{Spatial and Temporal Video Reasoning.} Recent work has improved the ability of VLMs to ground video reasoning in explicit spatial and temporal evidence. Temporal grounding methods such as Time-R1~\cite{wang2025time} and TVG-R1~\cite{wang2025time} use verifiable rewards to identify when relevant evidence appears. Spatial grounding methods such as SpaceR~\cite{ouyang2025spacer} use reinforcement learning to improve object-centric localization. Open-o3 Video~\cite{meng2025open} emits timestamped frames and bounding boxes as structured video evidence, while STVG-o1~\cite{gu2025thinking} uses box-level chain-of-thought for spatio-temporal grounding. These methods demonstrate the value of explicit evidence, but the inter-frame motion connecting localized observations remains implicit. Motion-o addresses this missing trajectory dimension by requiring the model to articulate how objects move between grounded observations.

\pipeline

\noindent \textbf{Evidence-based Video Reasoning.} A broader line of work studies explicit visual operations as intermediate reasoning steps, including detection, segmentation, tracking, and box-level grounding~\cite{fan2025grit,wang2025vgr,zhang2025deepeyes,gu2025thinking}. DeepEyes~\cite{zhang2025deepeyes} shows that reinforcement learning can incentivize image-tool reasoning, while TreeBench~\cite{wang2025treevgr} provides methodology for traceable box-level evidence in images. In video, OpenAI-o3~\cite{openai2025o3} popularizes thinking with images, and Open-o3 Video~\cite{meng2025open} extends this idea to thinking with frames. Motion-o follows this evidence-based direction, but changes the evidence unit. Instead of only grounding objects at isolated frames, it makes the motion relation between grounded observations explicit.


\textbf{Trajectory, Trace, and Action-Centric Representations.} A related line of work represents motion through trajectories, traces, or action-conditioned structures. In robotics, visual-trace methods such as LLARVA~\cite{llarva2025} use 2D trace prediction to align perception with action, while spatio-temporal grounding models such as VideoMolmo~\cite{videomolmo2025} improve localization and pointing over time. These works show that explicit traces are useful interfaces for dynamic visual tasks. Motion-o is complementary, but rather than treating trajectory outputs as action targets or localization outputs, it inserts motion into the VLM reasoning trace as an explicit variable connecting grounded observations to the final response.


\noindent \textbf{Reinforcement Learning-based training.} Reinforcement learning has recently become a central tool for improving multimodal reasoning. Video-R1~\cite{feng2025video} uses temporal-aware GRPO to improve video understanding, VideoChat-R1~\cite{li2025videochat} extends RL to spatio-temporal perception, Video-RTS~\cite{wang2025videorts} combines RL with test-time scaling, and DeepVideo-R1~\cite{park2025deepvideo} introduces difficulty-aware regularization for temporal structure. These methods show that RL can improve video reasoning, but primarily optimize answer quality. Motion-o instead uses RL to supervise an explicit motion variable, where the \texttt{<motion/>} tag must agree with the trajectory and remain dependent on temporal visual evidence.

Prior work has made substantial progress on temporal grounding, spatial grounding, and RL-based video reasoning. However, object motion is usually treated as an implicit latent dependency or an external control signal, rather than a verifiable statement inside the model's reasoning trace. Motion-o fills this gap by making trajectory information explicit and interpretable within grounded video reasoning.


\section{Introducing Motion-o}

\label{sec:method}
We present \textbf{Motion-o}, a framework that extends grounded video reasoning with an explicit trajectory reasoning step (Figure \ref{fig:hookah}). Motion-o augments spatio-temporal evidence chains with a discrete \texttt{<motion/>} descriptor that summarizes how an object moves between grounded observations.
Given a video-question pair $(\mathcal{V}, q)$, the model outputs
\[
y=\langle \texttt{think} \rangle R \langle /\texttt{think} \rangle \langle \texttt{answer} \rangle A \langle /\texttt{answer} \rangle,
\]
where $R$ contains (i) grounded spatio-temporal evidence and (ii) explicit motion descriptors. Motion-o is trained in two stages: (i) supervised fine-tuning (SFT) on motion-augmented reasoning traces that teach the model the structured format and syntax, and (ii) reinforcement learning that directly optimizes motion-centric rewards to encourage accurate and visually grounded motion descriptions.


Corresponding with Figure \ref{fig:hookah}, we first summarize the spatio-temporal evidence format that we extend to STT evidence (Sec.~\ref{sec:evidence}), then introduce MCoT and how we obtain supervision for motion tags (Sec.~\ref{sec:mcot}), and finally describe the training pipeline and trajectory-grounded rewards (Sec.~\ref{sec:training}).

\subsection{Spatial-Temporal-Trajectory (STT) Evidence Chains}
\label{sec:evidence}

Following Open-o3 Video~\cite{meng2025open}, we adopt a structured format for grounding reasoning in visual evidence. Given a video $\mathcal{V}$ and question $q$, the model produces a response organized as $\langle\text{think}\rangle R \langle/\text{think}\rangle \langle\text{answer}\rangle A \langle/\text{answer}\rangle$, where $R$ contains the reasoning trace and $A$ the final answer. Within $R$, each grounded claim is written as:
\begin{equation}
\langle\text{obj}\rangle o \langle/\text{obj}\rangle \langle\text{box}\rangle b \langle/\text{box}\rangle \;\text{at}\; \langle\text{t}\rangle t \langle/\text{t}\rangle\text{s},
\end{equation}
where $o$ is the object name, $b = [x_1, y_1, x_2, y_2]$ is a normalized bounding box, and $t$ is the timestamp in seconds. This representation creates an evidence chain that anchors language to concrete observations across time. For example, a model might write: ``The man is visible holding a glass at $47.5$s, and later at $54.2$s is smiling,'' each claim anchored by a precise spatial location and timestamp. However, even when the same object is referenced at multiple timestamps, the evidence chain remains a set of discrete snapshots: it specifies where the object was at $t_1$ and $t_2$, but leaves the trajectory dynamics between $t_1$ and $t_2$ implicit. Motion-o targets precisely this missing link.

\subsection{Motion Chain of Thought (MCoT)}
\label{sec:mcot}

We extend spatio-temporal evidence chains with explicit motion reasoning via a self-closing \texttt{<motion/>} tag to create STT evidence chains. The core idea is that after the model generates multiple temporal observations of the same object, it must summarize the motion that connects them. This converts a sequence of grounded boxes into a minimal, structured trajectory descriptor that is both interpretable and rewardable. 
\paragraph{Motion Tag Format}
Each \texttt{<motion/>} tag carries discrete attributes:
\textit{obj} (object name), \textit{dir} (direction: N, NE, E, SE, S, SW, W, NW, or STAT for stationary), \textit{speed} (stationary, slow, moderate, fast), and \textit{scale} (approaching, stable, receding). The format is:
\begin{equation}
\langle\texttt{motion}\;\text{obj}=\textit{name}\;\text{dir}=\textit{D}\;\text{speed}=\textit{S}\;\text{scale}=\textit{C}\;/\rangle.
\end{equation}

\textbf{Unified Reasoning Flow.} The complete reasoning chain then integrates temporal observations, spatial grounding, and motion reasoning into a single sequence. This enforces an explicit pathway: observe the object at multiple timestamps, ground each observation in a bounding box, emit a structured motion summary that links observations into a trajectory, and use that trajectory information to support the final answer. The \texttt{<motion/>} tag is a compact intermediate reasoning operator: it adds a small number of structured output tokens, but changes the semantics of the evidence chain by requiring the model to explicitly link grounded observations through a motion descriptor.

\textbf{Automatic Motion Computation.}
To teach MCoT without additional human annotation, we generate discrete motion primitives directly from the bounding boxes provided in each sample in the STGR dataset \cite{meng2025open}. Using the provided keyframe timestamps and object bounding boxes, we group bounding boxes by object identity across frames to obtain, for each object $o$, a time-ordered track $\mathcal{T}_o=\{(b_i,t_i)\}_{i=1}^{N}$, where $b_i=[x_1^i,y_1^i,x_2^i,y_2^i]$ denotes the box at time $t_i$. From $\mathcal{T}_o$, we compute box centroids $c_i$ and areas $a_i$, and use the induced displacement sequence $\Delta c_i = c_{i+1}-c_i$ to summarize motion. We convert each track into a discrete motion descriptor with three attributes: \textbf{direction}, computed by aggregating per-step displacement directions over the track with jitter-reducing weights and mapping the result to compass bins plus a stationary state; \textbf{speed}, computed from the average displacement rate normalized by object scale and discretized into ordinal bins; and \textbf{scale change}, computed from the relative change in bounding-box area over time and discretized as \{\texttt{approaching}, \texttt{stable}, \texttt{receding}\}.

\subsection{Training Pipeline}
\label{sec:training}

Our training follows the standard two-stage paradigm~\cite{meng2025open,feng2025video} with minimal modifications, while introducing Motion-o as a format-level extension that explicitly supervises and rewards trajectory transition. We first teach the model to emit well-formed motion descriptors as part of the reasoning trace through SFT and then directly optimize motion correctness and visual grounding through RL.

\textbf{Stage 1: Supervised Fine-Tuning.}
We fine-tune Qwen2.5-VL-7B~\cite{bai2025qwen} on the motion-augmented STGR data described above. Using standard cross-entropy optimization, the model learns to produce \texttt{<motion/>} tags as a structured continuation of spatio-temporal evidence chains: after grounding multiple observations of an object with \texttt{<obj><box><t>}, it emits a self-closing \texttt{<motion obj="..." dir="..." speed="..." scale="..."/>} that summarizes the trajectory connecting those observations. Beyond offline data augmentation, the only additional requirement in this stage is a system prompt instruction that specifies the tag schema and allowed attribute values. This stage establishes (i) strict format validity (\texttt{<think>/<answer>} and tag well-formedness), (ii) spatio-temporal grounding syntax, and (iii) the motion descriptor schema, ensuring the policy enters RL with stable, parseable trajectory reasoning traces.

\textbf{Stage 2: Reinforcement Learning.}
We build upon the training recipe in ~\cite{meng2025open}, adopting Group Sequence Policy Optimization (GSPO)~\cite{zheng2025group} to stably optimize long-form evidence-grounded video reasoning. Given a video--question prompt $x$, we sample a group of $K$ candidate completions 
$\{y^{(k)}\}_{k=1}^{K} \sim \pi_{\theta}(\cdot \mid x)$, evaluate each completion with a scalar reward $r(x,y^{(k)})$ and compute a group-normalized advantage. GSPO then updates the policy using sequence-level importance ratios, which better aligns optimization with sequence-level rewards and improves stability for long-horizon reasoning.

Following ~\cite{meng2025open}, we decompose the scalar reward into three components: 
\begin{equation}
r(x,y) = r_{\text{acc}}(x,y) + r_{\text{thk}}(x,y) + r_{\text{fmt}}(x,y).
\end{equation}
where, $r_{\text{acc}}$ encourages task-specific answer accuracy (e.g., exact match for MCQ, ROUGE for free-form QA), $r_{\text{fmt}}$ enforces strict structured output formatting in the reasoning chain (including well-formed \texttt{<motion/>} tags), and $r_{\text{thk}}$ incentivizes evidence-based reasoning. As in ~\cite{meng2025open}, we define the thinking reward as the sum of temporal and spatial grounding terms:

\begin{equation}
r_{\text{thk}}(x, y) \;=\; r_t(x, y) \;+\; r_s(x, y) \;+\; r_{\text{motion}}(x,y),
\label{eq:thinking_reward}
\end{equation}
where $r_t$ provides temporal supervision (with adaptive temporal proximity) and $r_s$ provides spatial supervision with temporal gating to ensure spatial rewards are only computed when predicted timestamps are sufficiently close to ground-truth.

\noindent\textbf{Motion-o extension: motion supervision inside $r_{\text{motion}}$.}
Our key extension is an explicit motion component
\begin{equation}
r_{\text{motion}}(x,y) \;=\; r_{\text{traj}}(x,y) \;+\; r_{\text{ground}}(x,y),
\end{equation}
which teaches the model to (i) describe trajectory dynamics consistently with underlying tracks, and (ii) rely on true temporal evidence rather than textual priors.


\noindent\textbf{Trajectory Reward.} For each object tracked across at least two temporal observations, we compute ground-truth motion bins from the bounding box trajectory and score the model's predicted \texttt{<motion/>} attributes via discrete bin matching with adjacency-aware partial credit:
\begin{equation}
r_{\text{traj}} = r_{\text{dir}} + r_{\text{speed}} + r_{\text{scale}}.
\end{equation}

Ground-truth bins are derived from centroid displacement and box area dynamics: direction from the dominant displacement vector quantized to eight compass points plus stationary, speed from displacement magnitude normalized by the object's bounding box diagonal, and scale from the log-ratio of box areas between the first and last observation. Trajectory scores are computed per object identity to ensure consistency over a single track.

\noindent\textbf{Visual Grounding Reward (Dual-Chain Verification).} To encourage the model to derive motion descriptions from actual visual observation rather than textual interpolation, we introduce a visual grounding reward via dual-chain verification (see Stage 2 inputs in Figure \ref{fig:hookah}). For each sample, we generate a second reasoning chain from a \textit{motion-masked} version of the video This is done by only freezing the sections of the video that correspond with  \texttt{<motion/>} tags, not the entire video. We then compare the \texttt{<motion/>} tags produced from the original video against those from the motion-masked input:
\begin{equation}
r_{\text{ground}} \;=\; \,\mathbbm{1}[d \neq d'] \;+\; \,\mathbbm{1}[s \neq s'] \;+\; \,\mathbbm{1}[c \neq c'],
\end{equation}
where $d, s, c$ and $d', s', c'$ denote the direction, speed, and scale attributes from the original and masked outputs, respectively. If predicted motion changes when motion evidence is removed, the model exhibits dependence on temporal cues ($r_{\text{ground}}\!\to\!1$); if predictions remain unchanged, it likely relies on non-visual shortcuts ($r_{\text{ground}}\!\to\!0$). Objects absent from the motion-masked output are treated as fully grounded.

\noindent\textbf{Synergy Between Trajectory and Visual Grounding Rewards.} Both the trajectory and visual grounding reward enforce complementary requirements. The trajectory reward enforces \emph{correctness} of motion descriptors with respect to ground-truth trajectories, while the visual grounding reward enforces \emph{evidence dependence} on temporal visual information rather than textual anticipation. Together they suppress two common failures: emitting \texttt{<motion/>} tags that are inconsistent with the underlying bounding box track, and emitting plausible \texttt{<motion/>} tags that could be produced without observing motion. This yields a self-reinforcing training signal: as the model learns to ground objects consistently across time, it is rewarded for emitting matching motion descriptors (higher $r_{\text{traj}}$); as it learns to rely on temporal evidence rather than priors, it is rewarded under motion masking (higher $r_{\text{ground}}$).



\subsection{Dataset Augmentation: Trajectory Grounding with Dense Motion Annotations} 
Existing spatio-temporal grounding datasets often provide bounding boxes at a small set of timestamps, yielding sparse trajectories that under-specify the motion connecting observations \cite{meng2025open}. We therefore construct a trajectory-grounding augmentation of the PerceptionLM (PLM) subset ~\cite{cho2025perceptionlm} by converting each sparse set of timestamped boxes into a temporally denser trajectory. For each tracked object, we preserve all original annotated keyframes and insert intermediate supervision points at a fixed temporal stride between adjacent annotated timestamps.  We preserve the original annotated keyframes and insert intermediate supervision points between adjacent timestamps using the dense mask annotations available in PLM. This produces a denser sequence of $(t,b_t)$ pairs for each tracked object, from which we derive the direction, speed, and scale-change descriptors described above. We release the augmented trajectories, derived motion descriptors, and splits to support reproducibility and to enable future work on explicit motion reasoning. The resource is intended primarily as a trajectory-grounding augmentation for evidence-based video reasoning. Further details on dataset splits and distribution are provided in the appendix in Section \ref{sec:dataset_info} and Figure \ref{fig:qual}.
\vstartable

\section{Experimental Results}
\label{sec:results}
We train Motion-o from Qwen2.5-VL-7B~\cite{bai2025qwen} following the two-stage pipeline in Sec.~\ref{sec:training}. To test whether MCoT provides additive gains on an already strong evidence-based model, we also add MCoT to the Open-o3 Video~\cite{meng2025open} checkpoint (denoted \textit{Open-o3 + MCoT}). Both variants are evaluated with and without the visual grounding reward $r_{\text{ground}}$. Following Open-o3 Video~\cite{meng2025open}, we report on the \textbf{V-STAR} benchmark~\cite{wang2023vstar} for spatio-temporal reasoning (What / When / Where, mAM, mLGM), and on \textbf{VideoMME}~\cite{fu2024videomme}, \textbf{WorldSense}~\cite{hong2025worldsense}, \textbf{TVGBench} \cite{wang2025time}, \textbf{MVBench}\cite{li2024mvbench}, and \textbf{MotionBench}\cite{hong2025motionbench} for general video understanding.

\subsection{Main Results}

\paragraph{Quantitative Analysis.} As reported in Table~\ref{tab:vstar} on the \textbf{V-STAR} benchmark, Open-o3 + MCoT achieves 36.6 mAM and 50.5 mLGM, improving over Open-o3 Video by +2.9 / +3.9 points. The gains are broad: What accuracy rises from 61.0 to 64.1, temporal grounding improves on both chains (Chain1: 24.5 $\rightarrow$ 27.3, Chain2: 24.0 $\rightarrow$ 26.8), and spatial grounding improves substantially (Chain1: 25.4 $\rightarrow$ 33.6, Chain2: 6.0 $\rightarrow$ 38.1), surpassing even the specialist Sa2VA-8B model on Where. Motion-o, trained from scratch on Qwen2.5-VL-7B, reaches 35.5 mAM, already surpassing Open-o3 Video and all open-source baselines. Removing $r_{\text{ground}}$ reduces mAM by 0.5 for Motion-o and 1.4 for Open-o3 + MCoT, with the largest drops on spatial grounding, confirming that dual-chain verification is critical for visually grounded motion descriptions. Beyond answer-level benchmark accuracy, we also evaluate whether the generated \texttt{<motion/>} tags match the ground truth, trajectory-derived labels in our augmented dataset (results in Appendix~\ref{sec:supp_motion_acc} and Table \ref{tab:motion_accuracy}). Motion-o produces tags that are aligned with the ground truth tracks, and removing the visual grounding reward reduces this alignment. This supports our claim that MCoT is not only a readable reasoning format, but also a measurable intermediate representation tied to trajectory evidence.

In Table~\ref{tab:main_results}),
Open-o3 + MCoT reaches 69.7 on VideoMME Overall and 41.5 on WorldSense, improving over Open-o3 Video by +6.1 and +4.0 points respectively while narrowing the gap with GPT-4o. Motion-o similarly improves over Qwen2.5-VL-7B across all metrics (+5.1 VideoMME, +3.1 WorldSense). These results confirm that explicit trajectory reasoning improves video understanding rather than competing with it.  Both Table~\ref{tab:vstar} and \ref{tab:main_results}  also compare Motion-o built on Qwen2.5 and Qwen3 backbones at a similar model scale. We observe that moving from Qwen2.5 to Qwen3 yields only modest gains, suggesting that at this scale the limiting factor is not only high-level reasoning ability, but also the ability to localize objects accurately. Because MCoT attributes are derived from changes in predicted object boxes, small localization errors can affect the estimated direction, speed, or scale even when the model captures the overall motion pattern. The qualitative examples in Figure~\ref{fig:supp1} support this interpretation. Future work would explore scaling Motion-o together with stronger VLM backbones and improved localization modules to produce more stable trajectory-grounded motion tags.

Additionally, MCoT increases output length because the model emits explicit motion descriptors in addition to spatio-temporal evidence. This overhead trades additional structured tokens for a rewardable motion statement. We quantify this trade-off by reporting token usage in Appendix~\ref{sec:supp_token_overhead}. 
\paragraph{Qualitative Analysis.} Figure~\ref{fig:qual} shows two representative outputs. In the first example, the model tracks Sheldon across three timestamps with varying camera angles and emits two \texttt{<motion/>} tags, both correctly identifying the subject as stationary despite significant viewpoint changes. The dense multi-point grounding enables the model to distinguish true stationarity from apparent visual displacement caused by camera cuts. Stationary tags make a falsifiable claim that an object's position and scale remain stable across the evidence window. Appendix~\ref{sec:supp_stationary} provides an additional example where a \texttt{STAT} tag directly supports the final answer by verifying that the relevant background object persists across time. In the second example, the model grounds a duck at four consecutive timestamps and summarizes its eastward trajectory with a single motion tag. Additional qualitative examples and analyses are provided in the appendix and supplementary. 
\resultstable
\motionqual
\subsection{Ablation Studies}
\label{sec:ablations}
\paragraph{Annotation Density.} As seen in the results in the top of 
\ablationdataformat
Table~\ref{tab:ablation_data_format}, dense annotations improve mAM from 31.2 to 35.5 (+4.3) and mLGM from 44.5 to 49.4 (+4.9), confirming that denser bounding box tracks yield substantially stronger trajectory supervision. VideoMME remains comparable (68.1 vs.\ 67.5), indicating that density primarily benefits motion-specific reasoning without harming general understanding.

\paragraph{Tag Representation and Zero-shot Prompting.}
The bottom of Table~\ref{tab:ablation_data_format} shows that replacing discrete bins with continuous numerical values (e.g., \texttt{speed="0.14 units/s", accel="0.1"}) causes a near-complete collapse: mAM drops from 35.5 to 2.1 and VideoMME from 67.5 to 14.4 (see Table~\ref{tab:ablation_data_format}, bottom).  The model fails to learn meaningful structure from continuous motion values, confirming that discrete ordinal bins are essential. The tags provide a tractable vocabulary that the language model can reliably produce, that aligns naturally with the bin-matching reward. We additionally evaluate zero-shot prompting with and without full motion-tag specification; details can be found in Appendix Section \ref{sec:supp_zeroshot} and Table \ref{tab:zeroshot_motion}. Structured prompting remains well below the fine-tuned Motion-o variants, indicating that explicit trajectory supervision and motion-grounded rewards are necessary for reliable and verifiable motion reasoning.

\section{Conclusion}
\label{sec:conclusion}
We introduced Motion-o, a lightweight framework that makes object motion explicit and verifiable in grounded video reasoning. By augmenting spatio-temporal evidence chains with MCoT \texttt{<motion/>} tags, Motion-o connects \emph{where} and \emph{when} objects appear with \emph{how} they move. Across video understanding benchmarks, this simple architecture-free extension improves motion-sensitive reasoning and shows that explicit trajectory variables can strengthen evidence-based VLM reasoning. Future work will extend MCoT beyond coarse single-object motion toward richer interaction, occlusion, camera-motion, and tracking-aware representations.

\definecolor{dirN}{HTML}{3B8BD4}
\definecolor{dirNE}{HTML}{5DCAA5}
\definecolor{dirE}{HTML}{EF9F27}
\definecolor{dirSE}{HTML}{ED93B1}
\definecolor{dirS}{HTML}{E24B4A}
\definecolor{dirSW}{HTML}{97C459}
\definecolor{dirW}{HTML}{AFA9EC}
\definecolor{dirNW}{HTML}{F5C4B3}
\definecolor{spdStat}{HTML}{888780}
\definecolor{spdSlow}{HTML}{3B8BD4}
\definecolor{spdMod}{HTML}{EF9F27}
\definecolor{spdFast}{HTML}{E24B4A}
\definecolor{sclStable}{HTML}{B4B2A9}
\definecolor{sclAppr}{HTML}{1D9E75}
\definecolor{sclRec}{HTML}{D85A30}
\definecolor{bestcolor}{HTML}{E8F5E9}
\renewcommand{\thetable}{S\arabic{table}}
\renewcommand{\thefigure}{S\arabic{figure}}

\clearpage
\bibliographystyle{plainnat}
\bibliography{main}

\appendix
\section*{Appendix}
This appendix provides additional details supporting the methodology, evaluation, and reproducibility of Motion-o. We first describe the training data, implementation setup, and compute resources used for the SFT and RL stages. We then report motion-tag accuracy on the augmented trajectory dataset, including zero-shot \texttt{<motion/>} prompting controls, and provide extended benchmark results and ablations. Finally, we document the prompt templates used for training, inference, and evaluation, and discuss broader impacts and responsible-use considerations.

\section{Training Details}
\label{sec:dataset_info}

\subsection{Training Data}
Both Motion-o and Open-o3 + MCoT models reported in this paper are trained using data derived from the
Spatio-Temporal Grounded Reasoning (STGR) dataset introduced in the Open-o3 paper~\cite{meng2025open}. Following that formulation, our training data is organized into SFT and  RL stages, with each sample associated with a source dataset and a task type.
The two models in this work, share the same RL data, but differ in their SFT mixtures to reflect different training priorities. 

\paragraph{Shared RL data.}
For both Motion-o and Open-o3 + MCoT, the RL stage uses the
same file, containing ~34k training samples. This shared RL set includes five task categories:
\textit{temporal-spatial free-form QA} (12,047), \textit{General video QA MCQ} (12,998),
\textit{visual QA} (5,000), \textit{temporal QA} (2,279), and \textit{general video QA free-form} (2,000). Using the same RL corpus for both models ensures that differences in final behavior are primarily attributable to the SFT data design rather than to different reinforcement learning supervision.

\begin{figure}
    \centering
    \includegraphics[width=0.43\textwidth]{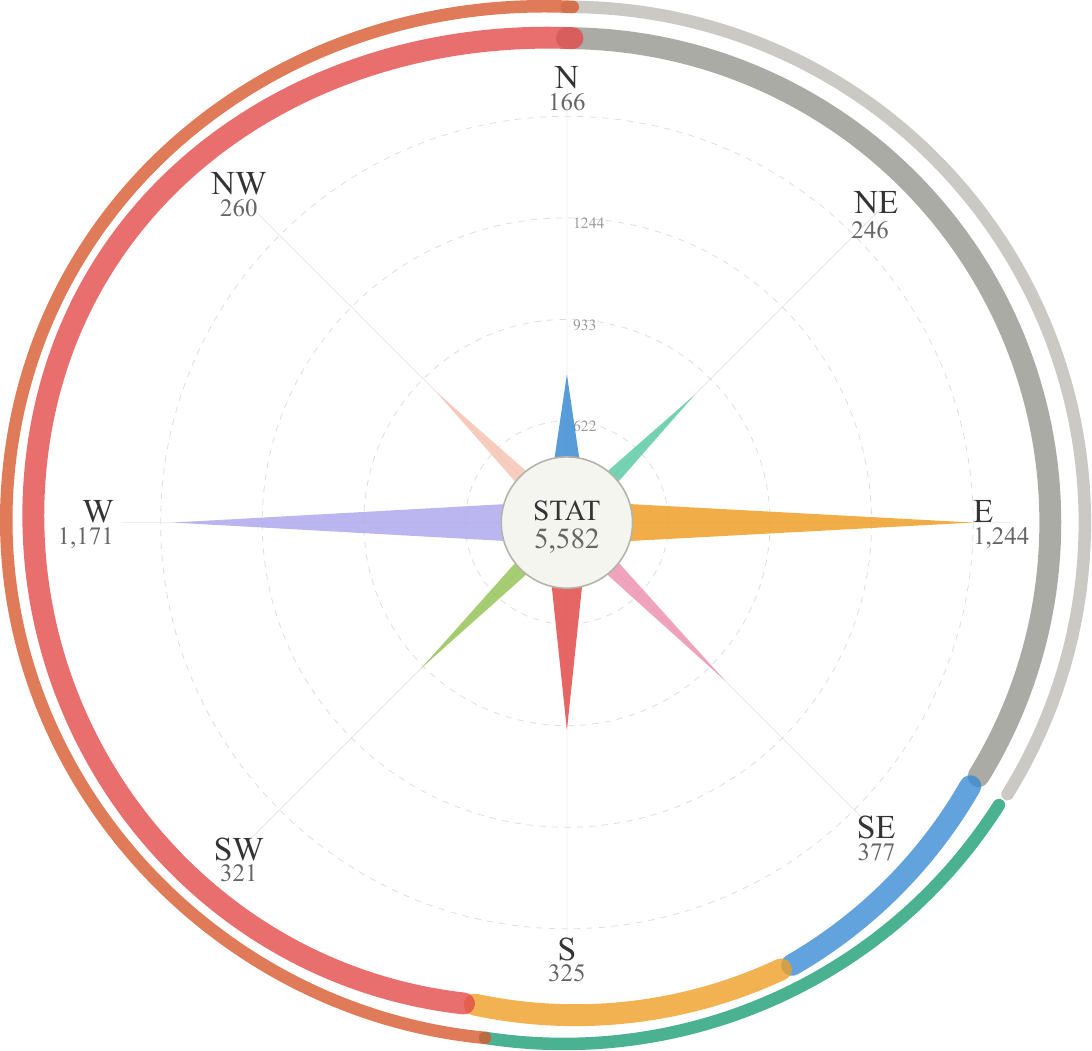}
    \caption{Motion data compass showing the distribution of observations across
    direction, speed, and scale. Directional wedges radiate from the center
    using square-root scaling:
    \textcolor{dirN}{\textbf{N}}~(166),
    \textcolor{dirNE}{\textbf{NE}}~(246),
    \textcolor{dirE}{\textbf{E}}~(1{,}244),
    \textcolor{dirSE}{\textbf{SE}}~(377),
    \textcolor{dirS}{\textbf{S}}~(325),
    \textcolor{dirSW}{\textbf{SW}}~(321),
    \textcolor{dirW}{\textbf{W}}~(1{,}171), and
    \textcolor{dirNW}{\textbf{NW}}~(260);
    the central circle is stationary~(5{,}582).
    The inner ring encodes speed:
    \textcolor{spdStat}{\textbf{stationary}}~(5{,}582),
    \textcolor{spdSlow}{\textbf{slow}}~(1{,}396),
    \textcolor{spdMod}{\textbf{moderate}}~(1{,}266),
    \textcolor{spdFast}{\textbf{fast}}~(1{,}448).
    The outer ring encodes scale:
    \textcolor{sclStable}{\textbf{stable}}~(5{,}751),
    \textcolor{sclAppr}{\textbf{approaching}}~(2{,}322),
    \textcolor{sclRec}{\textbf{receding}}~(1{,}619).
    Arc lengths are proportional to each category's share.}
    \label{fig:motion_compass}
\end{figure}

\paragraph{Motion-o SFT data.}
For Motion-o, the SFT stage uses a subset of the STGR dataset containing ~22k samples.
This mixture is broader and more balanced, combining motion-relevant grounded examples with a substantial amount of general video QA. Its task composition is: \textit{general video QA free-form} (2,000),  \textit{General video QA MCQ} (13,000), \textit{temporal QA} (491), and \textit{temporal-spatial free-form QA} (7,047). This design preserves broad video reasoning
coverage while still retaining a strong spatio-temporal grounding component, allowing Motion-o to couple explicit motion reasoning with more general video understanding capacity. Figure~\ref{fig:motion_compass} summarizes the distribution of motion labels in our augmented dataset. Examples are not dominated by a single direction: eastward and westward motions are closely matched, and the remaining directional bins are also populated across the compass. This indicates that the learned \texttt{<motion/>} tags are trained on a reasonably balanced set of trajectory orientations, rather than reflecting a strong directional skew. The inner and outer rings further show that the data covers a range of speed and scale-change categories, providing supervision over multiple facets of motion behavior. 

\paragraph{Open-o3 + MCoT SFT data.}
For Open-o3 + MCoT, we use the more motion-focused SFT mixture containing 7,838 samples. Its task composition is heavily concentrated in grounded motion-centric supervision:
\textit{temporal-spatial free-form QA} (6,849), \textit{general video QA MCQ} (817), \textit{general video QA free-form} (142), and \textit{temporal QA} (30). We use this more concentrated SFT mixture for the Open-o3 + MCoT variant because the Open-o3 base model has already been exposed to the broader STGR-style training recipe, including SFT and RL on this same corpus family, in its original training pipeline~\cite{meng2025open}.
Accordingly, for the MCoT variant we bias SFT more strongly toward motion-intensive grounded examples, with the goal of adapting the already grounded base model toward more explicit trajectory reasoning rather than repeating the full broad-coverage mixture used to establish its general grounded reasoning ability.

\paragraph{Key difference in training emphasis.}
The main distinction between the two models lies in the SFT stage. Motion-o is trained with a larger and more mixed SFT corpus that retains substantial general video QA supervision alongside grounded spatio-temporal data, whereas Open-o3 + MCoT uses a smaller but much more concentrated motion-oriented SFT set. Motion-o prioritizes a balance between broad video reasoning competence and explicit motion-aware grounding, while Open-o3 + MCoT prioritizes targeted adaptation of an already grounded base model toward trajectory-explicit reasoning. This also explains why Open-o3 + MCoT achieves higher overall results: the Open-o3 base model has already been trained on the STGR corpus~\cite{meng2025open}, giving it a strong prior over spatio-temporal grounding tasks. Our SFT stage therefore does not need to re-establish that general competence, and instead focuses narrowly on introducing \texttt{<motion/>} tag reasoning. The subsequent RL stage then builds directly on this foundation, using trajectory-consistency and visual grounding rewards to sharpen the model's ability to produce track-consistent, explicitly grounded motion claims. The performance advantage of Open-o3 + MCoT thus reflects this compounding of a well-initialized base with targeted motion supervision, rather than any fundamental architectural difference.

\subsection{Implementation Details}
\label{sec:implementation}
The SFT stage runs for 3 epochs on a single NVIDIA H200 GPU (140GB). The RL stage is conducted using 2 NVIDIA H200 GPUs (140GB).

\section{Motion Tag Accuracy on the Augmented Dataset}
\label{sec:supp_motion_acc}

Table~\ref{tab:motion_accuracy} evaluates whether the generated \texttt{<motion/>} tags are not only syntactically well-formed, but also aligned with the trajectory-derived motion labels in our augmented dataset. We compare predicted tags against ground-truth descriptors computed from dense bounding-box tracks and report accuracy for three attributes: direction (\textit{dir}), speed, and scale. For each attribute, we include both \textit{exact match}, where the predicted bin must equal the ground-truth bin, and \textit{adjacent match}, where predictions within one ordinal step are counted as correct. For direction, adjacent match corresponds to a tolerance of $\pm45^\circ$; for speed and scale, it corresponds to $\pm1$ rank. This relaxed metric is important because small localization errors or frame-level jitter can shift a descriptor into a neighboring bin even when the predicted motion is qualitatively reasonable.

The results show that Motion-o learns motion tags that are meaningfully aligned with dense trajectory evidence. In particular, adjacent-match accuracy is consistently higher than exact-match accuracy, indicating that many errors are near misses rather than completely incorrect motion interpretations. This is expected because the motion descriptors are derived from discretized bounding-box trajectories: small errors in object localization, centroid estimation, or box-area change can alter the final direction, speed, or scale bin. 

An important observation is that the improvement from the Qwen2.5-based variant to the Qwen3-based variant is modest. We interpret this as evidence that motion-tag accuracy is not limited only by the model's general language or reasoning capacity. Instead, a major bottleneck is the VLM's ability to localize the relevant object accurately and consistently across frames. The model may understand \emph{what} object is relevant and \emph{where} it approximately appears, but the derived motion tag depends on precise changes in the predicted box center and area over time. If the bounding boxes are spatially noisy, temporally inconsistent, or only coarsely aligned with the object, then the resulting direction, speed, and scale descriptors can be incorrect even when the model's semantic understanding of the event is broadly correct.

In the current pipeline, bounding boxes are generated by the VLM itself rather than by a dedicated detector or tracker. As a result, the system is constrained by the localization accuracy of the base model. This limitation is especially visible for attributes such as direction and scale, where small spatial errors can flip the discretized label. The ablation without the visual grounding reward further supports this interpretation. Removing $r_{\text{ground}}$ weakens the alignment between predicted motion tags and trajectory-derived labels, suggesting that the reward helps encourage motion claims that are more dependent on visual temporal evidence. However, the remaining errors indicate that reward-based verification cannot fully compensate for inaccurate localization. The quality of the motion reasoning trace therefore depends on both components: the model must learn to express motion explicitly, and the underlying visual grounding must provide sufficiently accurate object tracks for those expressions to be reliable.
\motionacc

\section{Zero-Shot \texttt{<motion/>} Prompting}
\label{sec:supp_zeroshot}

Table~\ref{tab:zeroshot_motion} evaluates whether the \texttt{<motion/>}
tag schema alone, without any fine-tuning, is sufficient to elicit
accurate motion reasoning. We prompt GPT-4o and Qwen2.5-VL-7B zero-shot
with and without the full tag specification. While structured prompting
yields modest gains over the unstructured baseline, the results fall
substantially short of Motion-o's fine-tuned performance
(Table~\ref{tab:motion_accuracy}), confirming that format compliance
alone does not substitute for trajectory-grounding supervision and the
visual grounding reward.

\zeroshottable
\tokentest

\section{Token Overhead of MCoT Responses}
\label{sec:supp_token_overhead}

MCoT makes motion reasoning explicit by adding structured \texttt{<motion/>} tags to the model's spatio-temporal evidence chain. This increases the length of generated responses compared to baselines that only produce an answer or a standard evidence trace. The results show that MCoT increases response length relative to the base VLM, as expected, because the model emits explicit spatio-temporal evidence and structured \texttt{<motion/>} descriptors rather than only a short final answer. However, this increase should be interpreted as the cost of making reasoning auditable. Motion-o processes more visual evidence and exposes the motion relation that would otherwise remain implicit.

Importantly, the trained MCoT variants remain token-efficient relative to strong grounded-reasoning baselines. Our MCoT-trained models produce responses in the range of 110-138 tokens on average, compared to 153 tokens for the Open-o3 base model. Thus, although Motion-o uses more tokens than a minimal answer-only baseline, it uses approximately 19\% fewer tokens than Open-o3 while providing explicit motion-grounded reasoning. This suggests that MCoT does not simply improve performance by making the model generate longer explanations; instead, trajectory supervision encourages more compact and structured reasoning traces.

We also observe that the visual grounding reward improves token efficiency. The variant without $r_{\text{ground}}$ produces substantially longer outputs, indicating that removing the grounding signal can lead to less concise reasoning. In contrast, the full Motion-o model learns to express the relevant trajectory information more directly through structured motion tags. 

\section{Extended Benchmark Results}
\label{sec:supp_extended}
\refertoio
Table~\ref{tab:main_results} reports results on three additional benchmarks: \textbf{TVGBench}~\cite{wang2025time} for temporal video grounding, \textbf{MVBench}~\cite{li2024mvbench} for multi-modal video understanding, and \textbf{MotionBench}~\cite{hong2025motionbench} (dev split) for motion-specific question answering. These complement the V-STAR, VideoMME, and WorldSense results in the main paper.

\paragraph{Quantitative Analysis} The benchmark results in Table ~\ref{tab:main_results} show a clear and consistent pattern: making motion explicit improves performance not only on motion-focused benchmarks, but also on broader video reasoning tasks that implicitly depend on dynamic understanding. Starting from the Qwen2.5-VL-7B baseline, Motion-o improves TVGBench from 16.3 to 37.6 (+21.3) and MotionBench from 35.0 to 60.0 (+25.0). When added on top of the Open-o3 base model, the Open-o3 + MCoT (Motion Chain-of-Thought) variant further raises TVGBench from 20.8 to 39.6 (+18.8), MVBench from 64.4 to 69.2 (+4.8), and MotionBench from 45.0 to 63.0 (+18.0). Notably, the best variant also surpasses the strongest open-source MVBench baseline in our table (69.2 vs.\ 67.9) and exceeds GPT-4o result on MotionBench by a wide margin (63.0 vs.\ 33.0). These gains suggest that explicit trajectory reasoning is improving a capability that many modern video benchmarks increasingly require. This pattern is important because the three benchmarks probe different surfaces of the same underlying problem. TVGBench emphasizes when evidence occurs and whether the model can temporally align its reasoning to the relevant segment. MotionBench is more directly motion-centric, explicitly testing categories such as action order, camera motion, location-related motion, and motion recognition. MVBench is broader and more heterogeneous, but many of its subsets also depend strongly on dynamics rather than static appearance alone, including action sequence, action prediction, and object interaction. Benchmarks are increasingly converging toward testing a model’s ability to reason about temporal change, transitions, and object evolution, rather than merely recognizing static content.

The key point we want to emphasize here is that benchmark construction has already started moving in a motion-centric direction, but most models have not. Many recent methods are designed as generic video reasoners: they may improve temporal grounding, spatial localization, or chain-of-thought style inference, but they still do not explicitly model the trajectory that connects observations. Existing evidence-based video reasoning frameworks have become increasingly strong at grounding  where an object is and when it appears, yet they typically leave how it moved between those observations implicit. This forces the model to interpolate dynamics internally, often relying on priors, textual smoothness, or weakly grounded heuristics rather than an explicit motion representation. Motion-o directly targets this gap by introducing with the structured \texttt{<motion/>} tag that summarizes direction, speed, and scale change between grounded observations, turning trajectory reasoning into an explicit and rewardable intermediate step. The gains do not come from architectural novelty, but from changing what the model is required to represent. Motion-o changes the supervision target and the reward structure so that motion is no longer an implicit byproduct of spatio-temporal grounding, but an explicit reasoning primitive. Prior models may include motion-sensitive metrics or temporal supervision, but the models themselves do not articulate motion patterns in their reasoning traces. In Motion-o, the trajectory is part of the reasoning trace itself, and the RL stage further encourages those motion claims to be both track-consistent and visually grounded under motion masking. 

\paragraph{Qualitative Analysis} Figure~\ref{fig:supp1} highlights a successful case where Motion-o captures a subtle but important temporal pattern: the baby moves in one circular direction and later reverses direction. In the generated reasoning trace, the motion grounding reflects this change explicitly, first marking the baby with westward motion and later with eastward motion. This example is qualitatively important because the key evidence is not merely the baby's presence or spatial location, but the change in trajectory over time. The model does not rely only on static snapshots; instead, it uses motion-grounded evidence across timestamps to support the answer. This illustrates the central motivation of Motion-o: making trajectory evolution explicit allows the model to reason about dynamic events that would otherwise remain implicit in standard spatio-temporal grounding pipelines. In this case, the model correctly tracks the temporal progression of the action and exposes the reversal through the structured motion tag, showing that the proposed motion-centric reasoning pathway can capture directional transitions that are essential for true video understanding.

Figure~\ref{fig:supp2} illustrates an important limitation of the current approach. Although the model arrives at the correct motion understanding the spatial localization is weak, as the predicted boxes do not tightly align with the referenced objects. This behavior is consistent with the design of the system. In the current pipeline, bounding boxes are produced by the underlying VLM backbone (here, Qwen), which is not a native object detector and is therefore not optimized for precise localization. As a result, Motion-o can still leverage coarse grounding and motion-aware reasoning to reach the correct answer, but the visual grounding itself may be spatially noisy. This suggests a natural direction for future improvement. One promising extension would be to integrate a stronger spatial localization component, such as a dedicated detector or grounding module, so that trajectory reasoning can operate on more accurate object tracks.

\paragraph{Stationary Motion as Informative Evidence.}
\label{sec:supp_stationary}
A natural concern with the \texttt{<motion/>} schema is that the \texttt{STAT}/\texttt{stationary}/\texttt{stable} configuration might behave as a trivial default; a label the model emits whenever it cannot detect movement, contributing little to downstream reasoning. Figure~\ref{fig:supp_stationary} shows that this is not the case in practice. The question asks what is visible in the background of the video \emph{when the miniature bottle is shown empty}, which requires the model to verify that a candidate object is present \emph{across} the relevant evidence window, not merely at a single keyframe. The model first grounds a hand near the bottle at $t{=}0$\,s and then grounds the seashells in the background at $t{=}20$\,s, and crucially
attaches a stationary motion tag to the seashells: \texttt{<motion obj="seashells" dir="STAT" speed="stationary" scale="stable"/>}. In the model's own reasoning trace, this tag is explicitly used to justify the claim that the seashells \emph{remain in
the frame}, which directly supports the final answer (\textbf{C.\ Shells}) over the distractors (water, cork, star).
 
This example highlights an asymmetric but important property of MCoT. Directional motion tags (e.g., \texttt{dir="W"} with \texttt{speed="moderate"}) make the model commit to a specific trajectory pattern between observations, and the trajectory and visual grounding rewards penalize disagreement with the underlying tracks. Stationary tags, by contrast, make the model commit to a different but equally falsifiable claim: that the object's position, velocity, and apparent scale are all approximately invariant across the evidence window. Because our trajectory reward is computed against
ground-truth bins derived from centroid displacement and box-area change (Sec.~\ref{sec:training}), an incorrectly emitted \texttt{STAT}/\texttt{stationary}/\texttt{stable} triple is penalized just like an incorrect directional triple. Conversely, when the underlying object truly is static, the dual-chain visual grounding
reward does not require the prediction to flip under motion masking, since freezing frames does not alter a genuinely stationary scene; the prediction is treated as fully grounded by construction. The stationary configuration therefore functions as a positive, rewardable assertion of temporal persistence rather than as an absence of information, which is why it can carry the weight of the answer in cases like Figure~\ref{fig:supp_stationary}. This complements the directional examples shown in the main paper (Figure~\ref{fig:qual}) and in Figure~\ref{fig:supp1}, and supports our broader claim that explicit motion descriptors, whether dynamic or static, convert implicit trajectory assumptions into verifiable evidence.

\suppstationary


\suppone

\supptwo

\lstdefinelanguage{PromptPy}{
  basicstyle=\ttfamily\footnotesize,
  keywordstyle=\color{promptkey}\bfseries,
  stringstyle=\color{promptstr},
  commentstyle=\color{promptcom}\itshape,
  morekeywords={role, content, type, image, video, text, user, system, assistant},
  morestring=[b]',
  morestring=[b]",
  morecomment=[l]{\#},
  showstringspaces=false,
  breaklines=true,
  breakatwhitespace=true,
  columns=fullflexible,
  keepspaces=true,
  literate=
    {<frame_1>}{{\textcolor{promptplaceholder}{\textit{<frame\_1>}}}}{9}
    {<frame_2>}{{\textcolor{promptplaceholder}{\textit{<frame\_2>}}}}{9}
    {<frame_n>}{{\textcolor{promptplaceholder}{\textit{<frame\_n>}}}}{9}
    {<video_frames>}{{\textcolor{promptplaceholder}{\textit{<video\_frames>}}}}{14}
    {<masked_video_frames>}{{\textcolor{promptplaceholder}{\textit{<masked\_video\_frames>}}}}{21}
    {<image>}{{\textcolor{promptplaceholder}{\textit{<image>}}}}{7}
    {<question_text>}{{\textcolor{promptplaceholder}{\textit{<question\_text>}}}}{17}
    {<options_text>}{{\textcolor{promptplaceholder}{\textit{<options\_text>}}}}{16}
    {<motion>}{{\textcolor{promptkey}{\textit{<motion/>}}}}{9}
}
 
\newtcblisting{promptbox}[2][]{
  listing only,
  listing options={language=PromptPy},
  colback=promptbg,
  colframe=promptframe,
  arc=2pt,
  outer arc=2pt,
  boxrule=0.6pt,
  left=6pt, right=6pt, top=4pt, bottom=4pt,
  title={\small\bfseries #2},
  fonttitle=\bfseries,
  breakable,
  enhanced,
  #1
}

\section{Prompt Templates}
\label{sec:prompts}
 
To support full reproducibility, this section documents the exact message
structures used during supervised fine-tuning (SFT), GSPO training, and
benchmark evaluation. All prompts follow the TRL~\cite{vonwerra2022trl}
standard chat format consumed by Qwen2.5-VL~\cite{bai2025qwen} and Qwen3-VL~\cite{bai2025qwen3}.
Placeholders of the form
\textcolor{promptplaceholder}{\textit{<...>}} are filled in per-example
at runtime, and the structured tag
\textcolor{promptkey}{\textit{<motion/>}} denotes the trajectory
descriptor introduced in Sec.~\ref{sec:mcot}.
 
\subsection{Frame Sampling}
 
For every training and evaluation example involving a video, we
uniformly sample $N$ frames from the video segment associated with the
question and pass them in temporal order, denoted
\textcolor{promptplaceholder}{\textit{<frame\_1>}}, \dots,
\textcolor{promptplaceholder}{\textit{<frame\_n>}}. We use $N{=}32$
frames during training and $N{=}64$ at evaluation unless otherwise
noted. The prompt structure is independent of $N$. For the dual-chain
visual grounding reward (Sec.~\ref{sec:training}), the
\emph{motion-masked} variant
\textcolor{promptplaceholder}{\textit{<masked\_video\_frames>}} replaces
intermediate frames with frozen copies of keyframes to remove temporal
motion cues while preserving appearance.
 
\subsection{Training Prompts (Task-Conditioned)}
\label{sec:supp_train_prompts}
 
Training samples are built with a task-conditioned \textit{system}
message followed by a multimodal \textit{user} message. The system
message specifies the output grammar that our reward parsers (format,
trajectory, and visual grounding) consume; we therefore list each
variant verbatim. Tasks correspond to the SFT/RL mixture described in
Appendix~\ref{sec:dataset_info}.
 
\subsubsection*{Temporal-Spatial Free-Form QA}
 
This is the primary trajectory-grounding task: the model must ground
each observation with \texttt{<obj>}, \texttt{<box>}, \texttt{<t>} tags
and emit a self-closing
\textcolor{promptkey}{\textit{<motion/>}} descriptor after the last
mention of any object that appears at $\geq 2$ timestamps.
 
\begin{promptbox}{SFT/GSPO Training Message --- Temporal-Spatial Free-Form QA}
messages = [
    {
        "role": "system",
        "content": [
            {
                "type": "text",
                "text": (
                    "A conversation between user and assistant. The user provides a video "
                    "and asks a question, and the assistant solves it by reasoning over "
                    "spatio-temporal evidence.\n"
                    "The reasoning process and answer are enclosed within "
                    "<think> </think> and <answer> </answer> tags respectively.\n"
                    "Inside <think>, every grounded observation must be written as "
                    "<obj>name</obj><box>[x1,y1,x2,y2]</box>at<t>seconds</t>s.\n"
                    "After the last grounded mention of an object that appears at >=2 "
                    "timestamps, emit a self-closing motion tag:\n"
                    "  <motion obj=\"name\" dir=\"D\" speed=\"S\" scale=\"C\"/>\n"
                    "where:\n"
                    "  dir   in {N, NE, E, SE, S, SW, W, NW, STAT}\n"
                    "  speed in {stationary, slow, moderate, fast}\n"
                    "  scale in {approaching, stable, receding}\n"
                    "The <answer> block contains plain text only "
                    "(no <obj>, <box>, or <t> tags)."
                ),
            },
        ],
    },
    {
        "role": "user",
        "content": [
            {"type": "video", "video": <video_frames>},
            {"type": "text",  "text": <question_text>},
        ],
    },
]
\end{promptbox}
 
\subsubsection*{General Video QA (MCQ)}
 
Multiple-choice general video QA reuses the
\textcolor{promptkey}{\textit{<motion/>}} schema inside
\texttt{<think>}, but constrains \texttt{<answer>} to a single option
letter.
 
\begin{promptbox}{SFT/GSPO Training Message --- General Video QA (MCQ)}
messages = [
    {
        "role": "system",
        "content": [
            {
                "type": "text",
                "text": (
                    "A conversation between user and assistant. The user provides a video, "
                    "a question, and a list of options. The assistant reasons inside "
                    "<think>...</think>, including <motion/> tags whenever motion is "
                    "relevant, and outputs only the option letter inside "
                    "<answer>...</answer>."
                ),
            },
        ],
    },
    {
        "role": "user",
        "content": [
            {"type": "video", "video": <video_frames>},
            {"type": "text",
             "text": "Question: <question_text>\nOptions:\n<options_text>"},
        ],
    },
]
\end{promptbox}
 
\subsubsection*{Temporal QA}
 
Temporal grounding samples ask the model to localize \emph{when} an
event occurs and constrain \texttt{<answer>} to a fixed time-range
format. No \textcolor{promptkey}{\textit{<motion/>}} tag is required.
 
\begin{promptbox}{SFT/GSPO Training Message --- Temporal QA}
messages = [
    {
        "role": "system",
        "content": [
            {
                "type": "text",
                "text": (
                    "A conversation between user and assistant. The user provides a video "
                    "and asks when an event occurs.\n"
                    "The assistant first thinks inside <think>...</think> and then "
                    "provides the final answer inside <answer>...</answer>.\n"
                    "The <answer> block must follow the format:\n"
                    "  From <t>t1</t>s to <t>t2</t>s"
                ),
            },
        ],
    },
    {
        "role": "user",
        "content": [
            {"type": "video", "video": <video_frames>},
            {"type": "text",  "text": <question_text>},
        ],
    },
]
\end{promptbox}
 
\subsubsection*{Visual QA (Image)}
 
Single-image grounding samples (used to maintain spatial-grounding
competence) require \texttt{<obj>}/\texttt{<box>} grounding inside
\texttt{<think>}, but no temporal or motion tags.
 
\begin{promptbox}{SFT/GSPO Training Message --- Visual QA (Image)}
messages = [
    {
        "role": "system",
        "content": [
            {
                "type": "text",
                "text": (
                    "A conversation between user and assistant. The user asks a question "
                    "about an image, and the assistant solves it.\n"
                    "The assistant first thinks inside <think>...</think> and then "
                    "provides the final answer inside <answer>...</answer>.\n"
                    "Inside <think>, every visual claim must be grounded as "
                    "<obj>name</obj><box>[x1,y1,x2,y2]</box>."
                ),
            },
        ],
    },
    {
        "role": "user",
        "content": [
            {"type": "image", "image": <image>},
            {"type": "text",  "text": <question_text>},
        ],
    },
]
\end{promptbox}
 
\subsection{Inference Prompt (Unified)}
\label{sec:supp_infer_prompts}
 
At inference time, our vLLM wrapper uses a single unified system prompt
(\texttt{MOTION\_SYSTEM\_PROMPT}) for all trajectory-grounded video
reasoning, regardless of whether the downstream task is free-form or
multiple-choice. This matches the temporal-spatial free-form QA training
distribution most closely and is also used by the dual-chain motion
masking pass: both the original and the motion-masked rollouts are
conditioned on the same system message, so any change in predicted
\textcolor{promptkey}{\textit{<motion/>}} attributes is attributable to
the temporal evidence rather than to prompt asymmetry.
 
\begin{promptbox}{Inference Message --- Unified Trajectory-Grounded Prompt}
messages = [
    {
        "role": "system",
        "content": [
            {
                "type": "text",
                "text": (
                    "A conversation between user and assistant. The user provides a video "
                    "and asks a question, and the assistant solves it.\n"
                    "The assistant first thinks about the reasoning process in the mind "
                    "and then provides the user with the answer.\n"
                    "Inside <think>...</think>, the assistant grounds each observation as "
                    "<obj>name</obj><box>[x1,y1,x2,y2]</box>at<t>t</t>s.\n"
                    "After grounding an object across multiple timestamps, the assistant "
                    "emits <motion obj=\"name\" dir=\"D\" speed=\"S\" scale=\"C\"/> "
                    "summarizing the trajectory, where:\n"
                    "  dir   in {N, NE, E, SE, S, SW, W, NW, STAT}\n"
                    "  speed in {stationary, slow, moderate, fast}\n"
                    "  scale in {approaching, stable, receding}\n"
                    "The reasoning process and answer are enclosed within "
                    "<think> </think> and <answer> </answer> tags respectively.\n"
                    "The answer part only requires a text response; tags like "
                    "<obj>, <box>, <t> are not needed."
                ),
            },
        ],
    },
    {
        "role": "user",
        "content": [
            {"type": "video", "video": <video_frames>},
            {"type": "text",  "text": <question_text>},
        ],
    },
]
\end{promptbox}
 
\subsection{Benchmark Evaluation Prompts}
\label{sec:supp_bench_prompts}
 
For benchmark evaluation, we follow each benchmark's official prompting
convention to ensure direct comparability with prior work. The
\textbf{V-STAR} benchmark explicitly evaluates spatio-temporal grounding
and reuses the unified inference prompt above. The remaining benchmarks
(MVBench, MotionBench, TVBench, VideoMME, WorldSense) constrain the
model to emit only a single option letter; the model is \emph{not}
required to produce \texttt{<obj>}/\texttt{<box>}/\texttt{<t>} or
\textcolor{promptkey}{\textit{<motion/>}} tags at the output level,
though it may still use them internally inside \texttt{<think>}.
 
\begin{promptbox}{Evaluation Message --- MVBench / MotionBench / TVBench (Letter-Only MCQ)}
messages = [
    {
        "role": "system",
        "content": [
            {
                "type": "text",
                "text": (
                    "Carefully watch the video and pay attention to every detail. "
                    "Based on your observations, select the option that best answers "
                    "the question. Answer with only the letter of your choice."
                ),
            },
        ],
    },
    {
        "role": "user",
        "content": [
            {"type": "video", "video": <video_frames>},
            {"type": "text",
             "text": "Question: <question_text>\nOptions:\n<options_text>"},
        ],
    },
]
\end{promptbox}
 
\subsection{Summary of Prompt Components}
 
\begin{table}[h]
\centering
\small
\setlength{\tabcolsep}{5pt}
\renewcommand{\arraystretch}{1.15}
\begin{tabular}{@{}p{0.20\linewidth} p{0.36\linewidth} p{0.36\linewidth}@{}}
\toprule
\textbf{Component} & \textbf{Training / Inference (Trajectory-Grounded)} & \textbf{Benchmark Evaluation (Letter-Only MCQ)} \\
\midrule
Frames & $N$ uniformly sampled, temporal order & $N$ uniformly sampled, temporal order \\
System message & Task-conditioned grammar (Sec.~\ref{sec:supp_train_prompts}) or unified inference prompt (Sec.~\ref{sec:supp_infer_prompts}) & ``Watch carefully \dots\ answer with only the letter'' \\
\texttt{<think>} requirements & \texttt{<obj>}/\texttt{<box>}/\texttt{<t>} grounding + \textit{<motion/>} tags & Free (not enforced at parse time) \\
\texttt{<answer>} content & Plain text, option letter, or \texttt{From <t>$t_1$</t>s to <t>$t_2$</t>s} (task-dependent) & Single option letter \\
Output target & Free-form generation with structured tags & Single letter $\in \{$A, B, C, D$\}$ \\
Scoring & $r_{\text{acc}} + r_{\text{thk}} + r_{\text{fmt}}$ (Sec.~\ref{sec:training}) & Exact-match top-1 accuracy \\
\bottomrule
\end{tabular}
\caption{Differences between trajectory-grounded prompts (used for SFT,
GSPO rollouts, and V-STAR/free-form inference) and letter-only
benchmark-evaluation prompts. The trajectory-grounded prompts define the
output grammar that Motion-o's reward parsers consume; the
benchmark-evaluation prompts follow each benchmark's official protocol
to ensure faithful comparison.}
\label{tab:prompt_components}
\end{table}

\section{Broader Impacts}
\label{sec:broader_impacts}

This work aims to get one step closer to improving the trustworthiness of VLMs by making motion-based reasoning more explicit, interpretable, and verifiable. By requiring models to expose trajectory-level evidence through structured \texttt{<motion/>} tags, Motion-o provides a mechanism for checking whether dynamic claims are supported by observed changes in the video rather than by static appearance cues or language priors. This direction may be especially valuable for domains where video understanding must be reliable, such as assistive technologies, robotics, healthcare monitoring, transportation, and other safety-critical settings.

A positive impact of this work is that explicit motion reasoning can make model outputs easier to audit. Instead of only producing a final answer, the model provides intermediate motion descriptors that can be compared against object tracks or perturbed inputs. This creates opportunities for verification and failure detection before a model response is trusted in downstream use. More broadly, methods that encourage models to verify their responses against visual evidence may support safer deployment of VLMs in settings where incorrect temporal or causal interpretations could have serious consequences.

At the same time, improved video understanding can also introduce risks. More reliable trajectory reasoning could be misused in privacy-sensitive surveillance, tracking, or behavioral analysis applications. Motion-o does not introduce new surveillance data or deploy models in such settings, but the underlying capability of more accurate motion interpretation should be handled carefully. We therefore view this work as primarily a step toward more transparent and accountable video reasoning, and we encourage future deployments to include appropriate privacy protections and task-specific validation before use in high-stakes environments.

\newpage
\end{document}

%% file: figs.tex
\newcommand{\comparison}{
\begin{figure}[t]
  \centering
  \includegraphics[width=0.92\columnwidth, trim={0 10pt 0 0},
    clip]{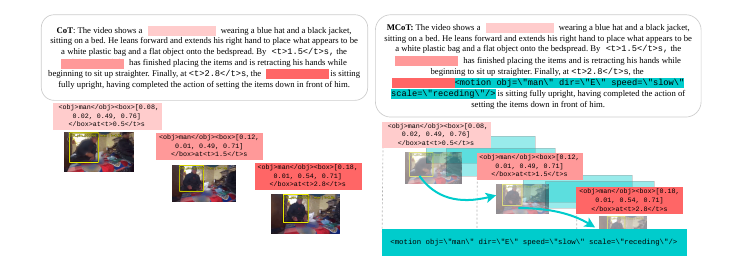}
  \caption{CoT vs. Motion Chain-of-Thought (MCoT). CoT in ~\cite{meng2025open} yields sparse temporal bounding boxes, forcing implicit inter-frame interpolation. MCoT adds an object-conditioned \texttt{<motion/>} tag that parameterizes the dynamics between observations.}
  \label{fig:comparison}
  \vspace{-10pt}
\end{figure}
}

\newcommand{\pipeline}{
\begin{figure}[t]
\centering
\includegraphics[width=0.8\columnwidth]{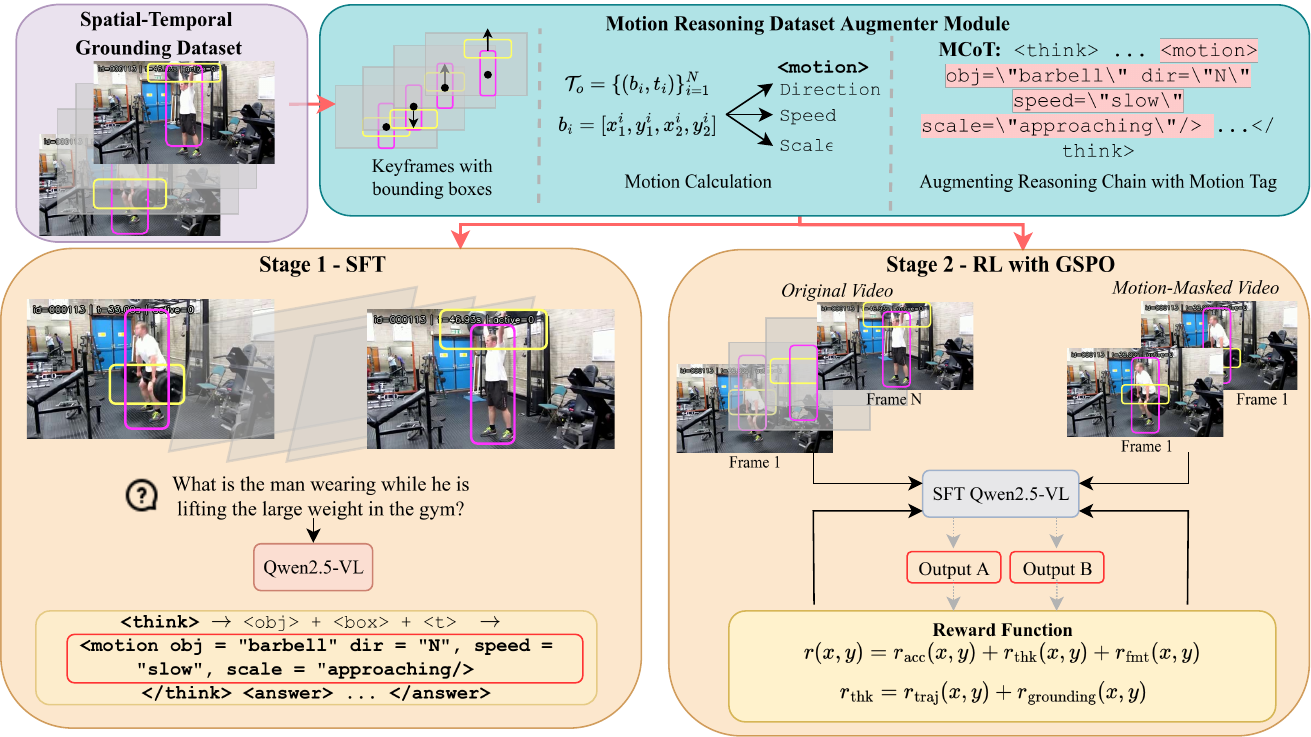}
\caption{Motion-o pipeline. Starting from spatio-temporal grounding data~\cite{meng2025open} and our dense interpolated boxes, we derive motion primitives (direction, speed, scale) and insert them as MCoT \texttt{<motion/>} tags. Motion-o is then trained with SFT (Stage 1) to learn the STT/MCoT format and RL (GSPO, Stage 2) to reward trajectory-consistent tags that change when temporal evidence is removed.}
\label{fig:hookah}
\end{figure}
}

\newcommand{\motionqual}{
\begin{figure}[t]
\centering
\includegraphics[width=\columnwidth, trim={0 120pt 0 0},
    clip]{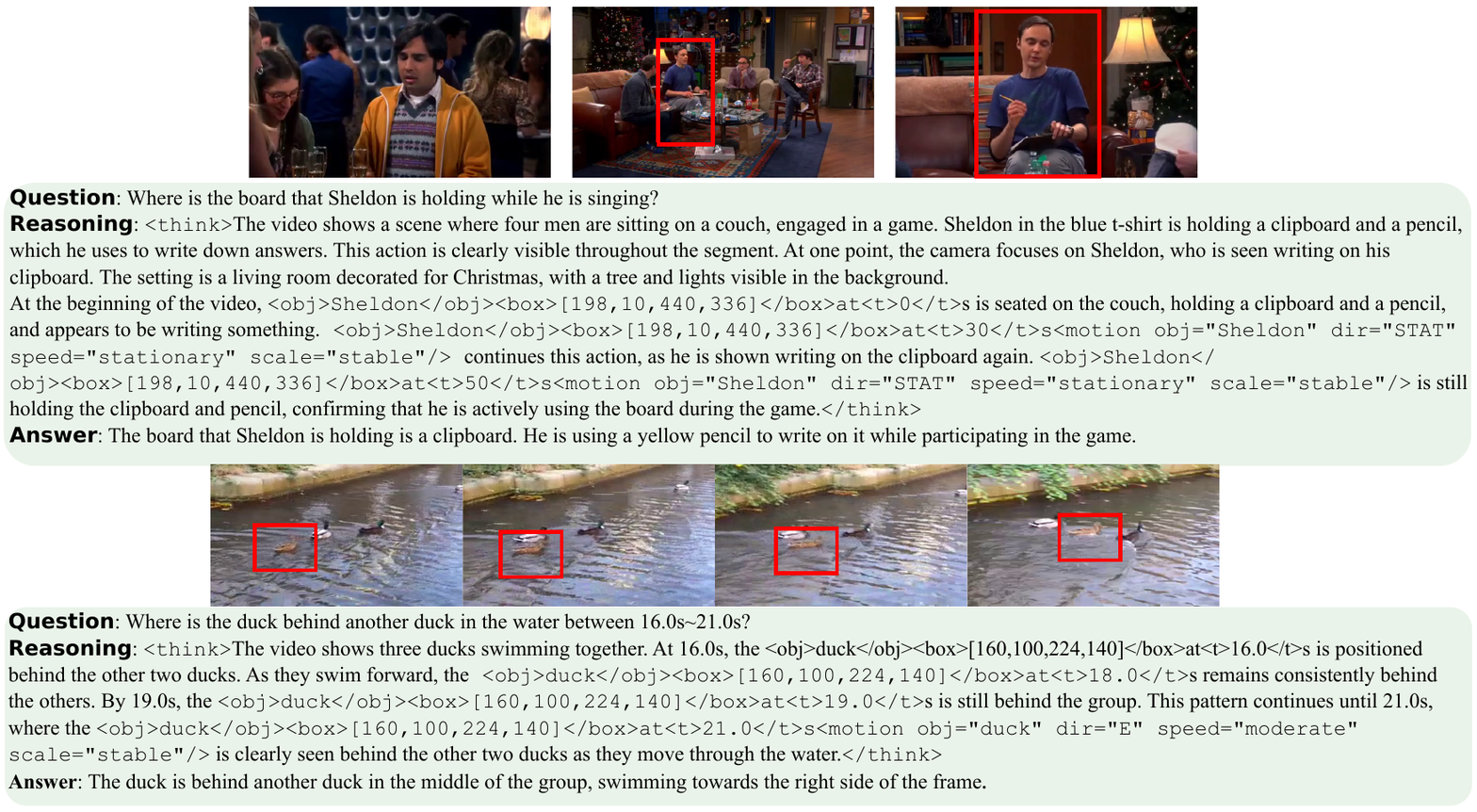}
\caption{Qualitative examples of Motion-o reasoning. \textbf{Top:} Motion-o tracks Sheldon across camera changes and correctly emits stationary \texttt{<motion/>} tags, distinguishing true stationarity from apparent displacement. 
\textbf{Bottom:} Motion-o grounds a swimming duck across timestamps and summarizes its eastward trajectory with a single moderate-speed motion tag.}
\label{fig:qual}
\vspace{-15pt}
\end{figure}
}

\newcommand{\suppone}{%
\begin{figure}[t]
\centering
\includegraphics[width=\columnwidth]{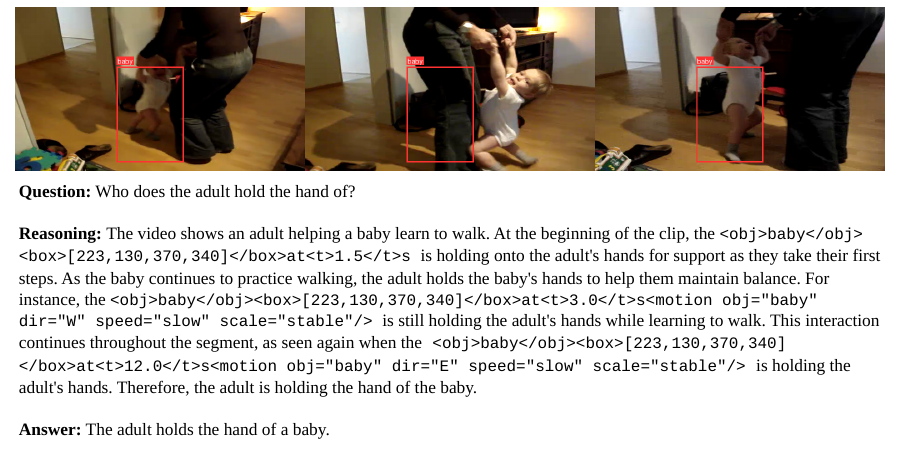}
\caption{Qualitative example showing successful motion-centric reasoning:
the model captures the baby's change in movement direction over time
through explicit motion grounding.}
\label{fig:supp1}
\vspace{-17pt}
\end{figure}
}
 
\newcommand{\supptwo}{%
\begin{figure}[t]
\centering
\includegraphics[width=0.8\columnwidth]{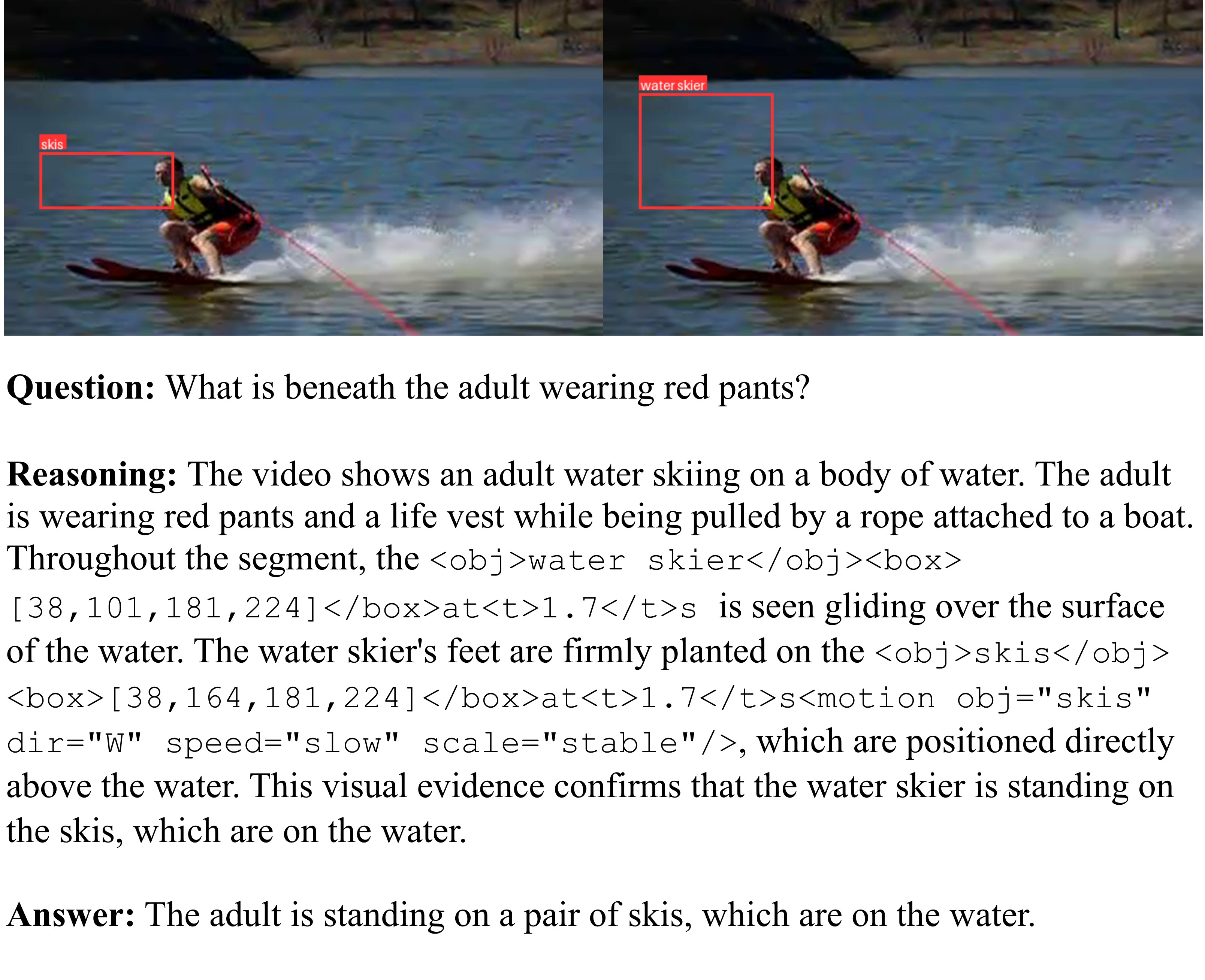}
\caption{Qualitative limitation example: the model reaches the correct
answer, but the spatial grounding is imprecise, highlighting the need
for stronger localization modules.}
\label{fig:supp2}
\vspace{-17pt}
\end{figure}
}
 
\newcommand{\suppstationary}{%
\begin{figure}[t]
\centering
\includegraphics[width=0.95\columnwidth]{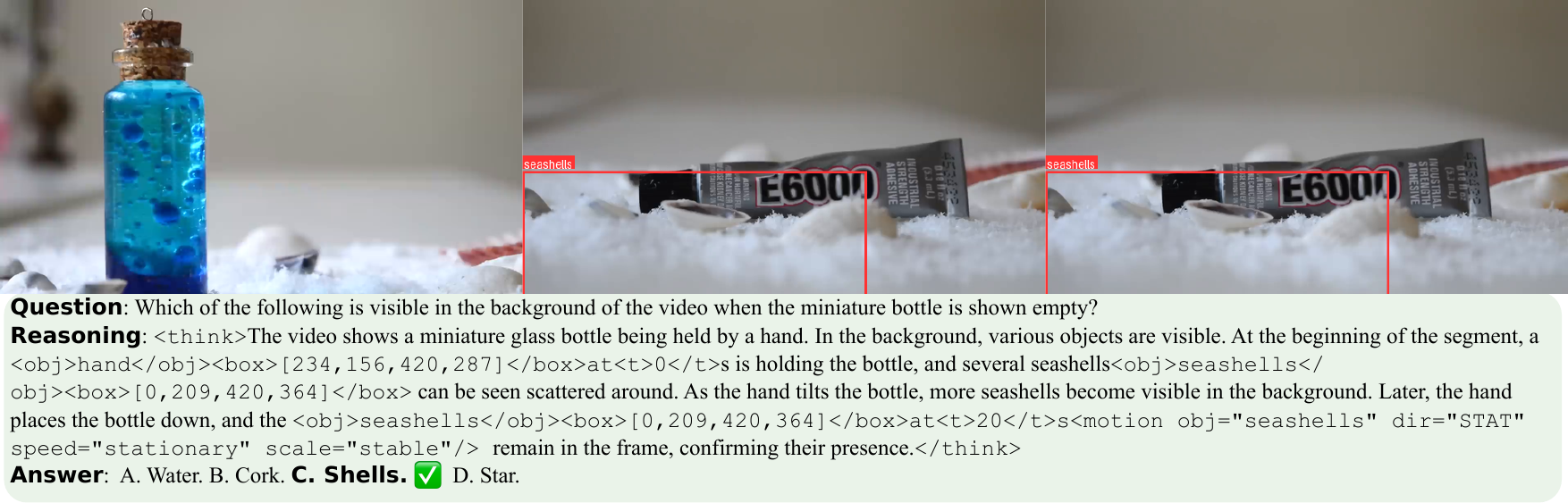}
\caption{\textbf{Stationary motion as informative evidence.} Even when an
object does not move, the explicit \texttt{<motion/>} tag carries
non-trivial reasoning value. Here, the model grounds the seashells in
the background and emits
\texttt{<motion obj="seashells" dir="STAT" speed="stationary"
scale="stable"/>}, certifying that the seashells \emph{remain in the
frame} across the evidence window rather than appearing only briefly.
This persistence claim is precisely what the question requires
(``visible in the background \dots\ when the miniature bottle is shown
empty''), and it allows the model to confidently select
\textbf{C.\ Shells} over the distractors. The example illustrates that
\texttt{dir="STAT"} is not a degenerate or empty label: it is an
explicit, rewardable assertion of temporal persistence that complements
the directional motion cases shown in the main paper.}
\label{fig:supp_stationary}
\vspace{-10pt}
\end{figure}
}

\newcommand{\refertoio}{
\begin{figure}[t]
\centering
\includegraphics[width=0.95\columnwidth]{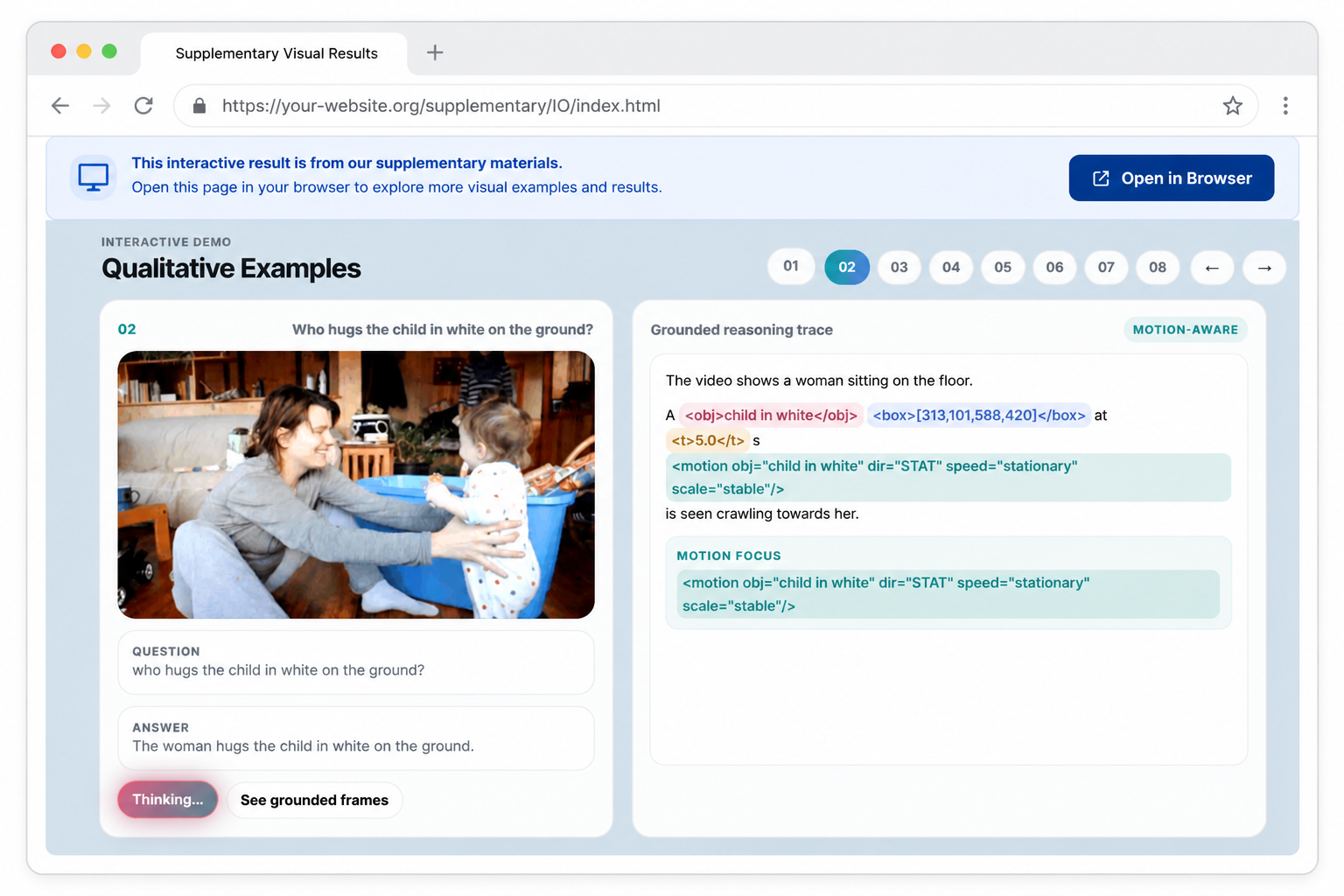}
\caption{
\textbf{Additional interactive qualitative results.}
We provide an interactive HTML in the supplementary materials that allows you to browse additional visual examples, grounded reasoning traces, motion-aware annotations, and grounded frames. 
Please open \texttt{supplementary/index.html} in a web browser to view the full set of qualitative results beyond the examples shown in the paper.}
\label{fig:seeio}
\vspace{-10pt}
\end{figure}
}

\newcommand{\tokentest}{
\begin{figure}[t]
\centering
\includegraphics[width=0.90\columnwidth]{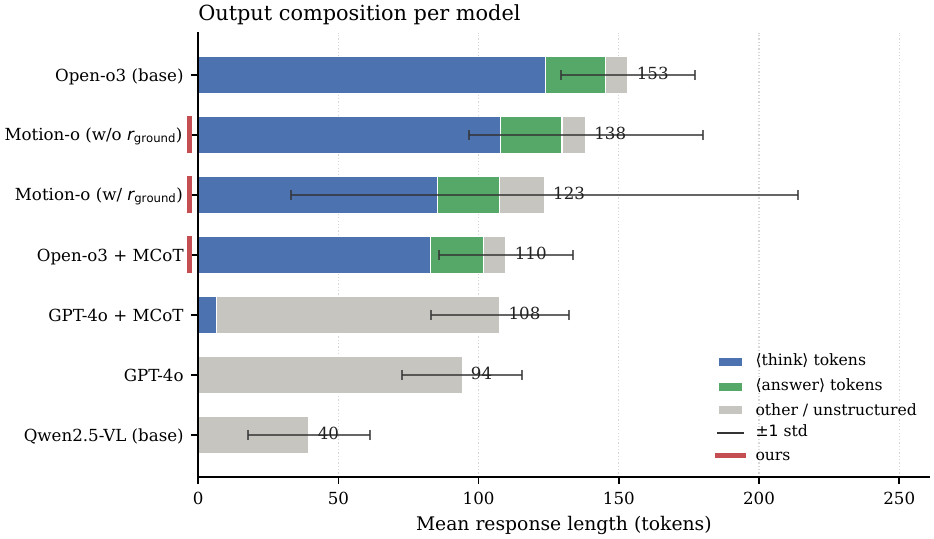}
\caption{\textbf{Generated-token overhead across model variants.}
We report the mean response length for each model (ran over 100 samples), decomposed into \texttt{<think>} tokens, \texttt{<answer>} tokens, and other unstructured tokens, with error bars showing $\pm 1$ standard deviation. Although Motion-o introduces explicit structured tokens for motion grounding, both Motion-o variants and Open-o3 + MCoT remain shorter than Open-o3 base — Motion-o (w/ $r_{\mathrm{ground}}$) uses \textbf{19\% fewer tokens} (123 vs.\ 153) and Open-o3 + MCoT uses \textbf{28\% fewer} (110 vs.\ 153) — while providing rewardable trajectory evidence, indicating that MCoT supervision densifies rather than lengthens the reasoning trace.}
\label{fig:token_overhead}
\vspace{-10pt}
\end{figure}
}

%% file: tabs.tex
\newcommand{\vstartable}{
\begin{table*}[t]
  \centering
  \caption{Performance on the \textbf{V-STAR} benchmark, which evaluates spatio-temporal reasoning across three dimensions. Chain1 denotes \textit{what--when--where}; Chain2 corresponds to \textit{what--where--when}. mAM is the average of arithmetic mean, and mLGM is the average of modified logarithmic geometric mean, combining temporal and spatial alignment. * indicates re-evaluation using the vLLM framework with 16 sampled frames. The $r_{\text{ground}}$ column indicates whether the visual grounding reward is enabled (\rgon) or disabled (\rgoff) in our methods. \textbf{Bold} numbers denote the best results, while \underline{underlined} numbers indicate the second best.}
  \label{tab:vstar}
  \setlength{\tabcolsep}{5pt}
  \renewcommand{\arraystretch}{1.15}
  \footnotesize
  \resizebox{0.96\textwidth}{!}{%
  \begin{tabular}{l c c cc cc c c}
    \toprule
    \multirow{2}{*}{Model}
      & \multirow{2}{*}{$r_{\text{ground}}$}
      & What
      & \multicolumn{2}{c}{When (Temporal IoU)}
      & \multicolumn{2}{c}{Where (Visual IoU)}
      & \multicolumn{2}{c}{Overall} \\
    \cmidrule(lr){3-3}\cmidrule(lr){4-5}\cmidrule(lr){6-7}\cmidrule(lr){8-9}
      &
      & Acc
      & Chain1 & Chain2
      & Chain1 & Chain2
      & mAM & mLGM \\
    \midrule
    GPT-4o               & & 60.8 & 16.7 & 12.8 & 6.5  & 3.0  & 26.8 & 38.2 \\
    Gemini-2-Flash       & & 53.0 & 24.5 & 23.8 & 4.6  & 2.2  & 26.9 & 35.6 \\
    \midrule
    Video-LLaMA3         & & 41.9 & 23.0 & 23.1 & 0.9  & 0.2  & 21.7 & 27.0 \\
    LLaVA-Video          & & 49.5 & 10.5 & 12.2 & 1.9  & 1.3  & 20.8 & 27.3 \\
    VideoChat2           & & 36.2 & 13.7 & 12.5 & 2.5  & 1.0  & 17.0 & 20.3 \\
    Oryx-1.5-7B          & & 20.5 & 13.5 & 14.8 & 10.1 & 3.5  & 15.1 & 13.8 \\
    InternVL-2.5-8B      & & 44.2 & 8.7  & 7.8  & 0.7  & 0.1  & 17.6 & 24.9 \\
    Qwen2.5-VL-7B* (base)& & 33.5 & 15.4 & 13.8 & 17.0 & 2.5  & 19.3 & 22.4 \\
    Qwen3-VL-8B* (base)  & & 36.0 & 23.5 & 13.9 & 8.5 & 6.8 & 20.2 & 23.1 \\

    \midrule
    TRACE                & & 17.6 & 19.1 & 17.1 & 0.0  & 0.0  & 12.0 & 13.3 \\
    Sa2VA-8B             & & 16.4 & 0.1  & 0.0  & \underline{32.3} & \underline{37.5} & 17.1 & 20.3 \\
    \midrule
    Open-o3 Video        & & 61.0 & 24.5 & 24.0 & 25.4 & 6.0  & 33.7 & 46.6 \\
    \midrule
    Motion-o (Qwen2.5-VL-7B) & \rgoff & 62.2 & 26.2 & 25.8 & 26.6 & 12.3 & 35.0 & 48.8 \\
    Motion-o (Qwen2.5-VL-7B) & \rgon  & 62.7 & 26.6 & 26.2 & 28.2 & 15.1 & 35.5 & 49.4 \\
    Motion-o (Qwen3-VL-8B) & \rgoff & 63.1 & 26.8 & 26.4 & 29.5 & 16.8 & 35.7 & 49.6 \\
    Motion-o (Qwen3-VL-8B) & \rgon  & \underline{63.6} & \underline{27.0} & \underline{26.7} & 31.2 & 18.4 & \underline{36.1} & \underline{50.1} \\
    Open-o3 + MCoT           & \rgoff & 63.4 & 26.3 & 25.6 & 30.5 & 31.4 & 35.2 & 49.1 \\
    \textbf{Open-o3 + MCoT}  & \rgon  & \textbf{64.1} & \textbf{27.3} & \textbf{26.8} & \textbf{33.6} & \textbf{38.1} & \textbf{36.6} & \textbf{50.5} \\
    \bottomrule
  \end{tabular}
  }
   \vspace{-15pt}
\end{table*}
}

\newcommand{\resultstable}{
\begin{table*}[t]
  \centering
  \caption{Performance across video understanding, temporal grounding, and motion-centric benchmarks. The $r_{\text{ground}}$ column indicates whether the visual grounding reward is enabled (\rgon) or disabled (\rgoff) in our methods. \textbf{Bold} numbers denote the best results, while \underline{underlined} numbers indicate the second best.}
  \label{tab:main_results}
  \setlength{\tabcolsep}{5pt}
  \renewcommand{\arraystretch}{1.15}
  \footnotesize
  \resizebox{\textwidth}{!}{%
  \begin{tabular}{l c cc cc c c c}
    \toprule
    \multirow{2}{*}{Model}
      & \multirow{2}{*}{$r_{\text{ground}}$}
      & \multicolumn{2}{c}{VideoMME}
      & \multicolumn{2}{c}{WorldSense}
      & \multirow{2}{*}{TVGBench}
      & \multirow{2}{*}{MVBench}
      & \multirow{2}{*}{MotionBench} \\
    \cmidrule(lr){3-4}\cmidrule(lr){5-6}
      &
      & Overall & Long
      & Overall & Recognition
      & & & \\
    \midrule
    GPT-4o                & & 71.9 & -    & \textbf{42.6} & -    & 26.8    & 43.50    & 33.0 \\
    Gemini-2.5-Pro        & & 84.8 & -    & 65.1 & -    & 30.3 & 45.5 & 66.3 \\
    Gemini-3.0            & & \textbf{86.9} & -    & -    & -    & 33.6    & 65.5 & 70.4 \\
    \midrule
    Qwen2.5-VL-7B (base)  & & 62.4 & 50.8 & 36.1 & 33.7 & 16.3 & 66.9 & 35.0 \\
    Qwen3-VL-8B (base) & & 64.1 & 52.3 & 37.2 & 35.1 & 18.4 & 69.0 & 38.0 \\
    VideoRFT-7B           & & 59.8 & 50.7 & 38.2 & 36.6 & 14.3 & 61.4 & 36.2 \\
    VideoR1-7B            & & 61.4 & 50.6 & 35.5 & 32.8 & 9.6  & 62.7 & 38.5 \\
    Open-o3 Video         & & 63.6 & 54.9 & 37.5 & 36.8 & 20.8 & 64.4 & 45.0 \\
    \midrule
    Motion-o (Qwen2.5-VL-7B) & \rgoff & 64.2 & 55.5 & 37.8 & 37.1 & 35.6 & 65.2 & 55.0 \\
    Motion-o (Qwen2.5-VL-7B) & \rgon  & 67.5 & 58.6 & 39.2 & 38.5 & 37.6 & 65.8 & 60.0 \\
    Motion-o (Qwen3-VL-8B) & \rgoff & 64.8 & 56.1 & 38.4 & 37.7 & 36.8 & 66.1 & 57.0 \\
    Motion-o (Qwen3-VL-8B) & \rgon  & 68.2 & 59.3 & 40.1 & 39.4 & 38.9 & 67.0 & 62.0 \\
    Open-o3 + MCoT           & \rgoff & 66.3 & 57.3 & 38.9 & 38.2 & 38.2 & 67.2 & 60.0 \\
    \textbf{Open-o3 + MCoT}  & \rgon  & \underline{69.7} & \textbf{60.3} & \underline{41.5} & \textbf{40.8} & \textbf{39.6} & \textbf{69.2} & \textbf{63.0} \\
    \bottomrule
  \end{tabular}%
  }
   \vspace{-10pt}
\end{table*}
}

\newcommand{\ablationdataformat}{
\begin{wraptable}{r}{0.48\textwidth}
  \vspace{-5pt}
  \centering
  \caption{Ablation on annotation density and tag format. Dense/Discrete is our default.}
  \label{tab:ablation_data_format}
  \setlength{\tabcolsep}{3pt}
  \renewcommand{\arraystretch}{0.82}
  \scriptsize
  \begin{tabular}{llccc}
    \toprule
    Abl. & Config & mAM & mLGM & VMME \\
    \midrule
    \multirow{2}{*}{Dens.}
      & Sparse/Disc. & 31.2 & 44.5 & \textbf{68.1} \\
      & Dense/Disc.  & \textbf{35.5} & \textbf{49.4} & 67.5 \\
    \midrule
    \multirow{2}{*}{Tag}
      & Dense/Num.   & 2.1 & 5.4 & 14.4 \\
      & Dense/Disc.  & \textbf{35.5} & \textbf{49.4} & \textbf{67.5} \\
    \bottomrule
  \end{tabular}
  \vspace{-10pt}
\end{wraptable}
}

\newcommand{\motionacc}{
\begin{table}[t]
\centering
\caption{\textbf{\texttt{<motion/>} tag accuracy} on the augmented trajectory-grounding dataset. We report exact-match accuracy (\%) for each motion attribute. ``Avg'' denotes the macro-average over three attributes. The $r_{\text{ground}}$ column indicates whether the visual grounding reward is enabled (\rgon) or disabled (\rgoff). Rows marked \textit{SFT only} use the cold-start supervised stage without reinforcement learning.}
\label{tab:motion_accuracy}
\setlength{\tabcolsep}{4.5pt}
\renewcommand{\arraystretch}{1.15}
\begin{tabular}{l c cccc}
\toprule
\textbf{Model} & $r_{\text{ground}}$ & \textbf{Direction} & \textbf{Speed} & \textbf{Scale} & \textbf{Avg} \\
\midrule
\multicolumn{6}{l}{\textit{\small SFT cold-start only (no RL)}} \\
\midrule
Motion-o (Qwen2.5-VL-7B)        & & 51.3 & 68.2 & 71.4 & 63.6 \\
Motion-o (Qwen3-VL-8B)          & & 52.8 & 69.7 & 72.6 & 65.0 \\
Open-o3 + MCoT                  & & 68.1 & 79.4 & 80.2 & 75.9 \\
\midrule
\multicolumn{6}{l}{\textit{\small SFT + GRPO (full pipeline)}} \\
\midrule
Motion-o (Qwen2.5-VL-7B)        & \rgoff & 72.1 & 81.4 & 82.7 & 78.7 \\
Motion-o (Qwen2.5-VL-7B)        & \rgon  & 78.9 & 88.3 & 88.4 & 85.2 \\
Motion-o (Qwen3-VL-8B)          & \rgoff & 73.5 & 83.1 & 84.2 & 80.3 \\
Motion-o (Qwen3-VL-8B)          & \rgon  & 79.6 & 89.1 & 89.3 & 86.0 \\
Open-o3 + MCoT                  & \rgoff & 74.8 & 84.2 & 85.6 & 81.5 \\
\rowcolor{bestcolor}
\textbf{Open-o3 + MCoT}         & \rgon  & \textbf{80.4} & \textbf{90.7} & \textbf{90.1} & \textbf{87.1} \\
\bottomrule
\end{tabular}
\end{table}
}

\newcommand{\zeroshottable}{
\begin{table}[t]
\centering
\caption{\textbf{Zero-shot \texttt{<motion/>} prompting} evaluated on motion tag accuracy (\%) (using the MCoT augmented subset of data) and V-STAR (10\% of the dataset). ``w/ prompt'' denotes zero-shot inference with the full \texttt{<motion/>} schema; ``w/o prompt'' is the standard baseline. No fine-tuning is applied.}
\label{tab:zeroshot_motion}
\setlength{\tabcolsep}{5pt}
\renewcommand{\arraystretch}{1.15}
\begin{tabular}{l c cccc ccc}
\toprule
\multirow{2}{*}{\textbf{Model}} & \multirow{2}{*}{\textbf{Prompt}}
  & \multicolumn{4}{c}{\textbf{Motion Tag Acc (\%)}}
  & \multicolumn{3}{c}{\textbf{V-STAR}} \\
\cmidrule(lr){3-6}\cmidrule(lr){7-9}
& & Dir & Speed & Scale & Avg & Temp tIoU & R@0.3 & Spatial mIoU \\
\midrule
GPT-4o & w/o \texttt{<motion/>} &  0.0 &  0.0 &  0.0 &  0.0 & 23.6 & 28.0 & 10.3 \\
GPT-4o & w/  \texttt{<motion/>} & \textbf{31.9} & \textbf{29.8} & \textbf{30.5} & \textbf{30.7} & \textbf{25.4} & \textbf{29.0} & \textbf{12.5} \\
\midrule
Qwen2.5-VL-7B & w/o \texttt{<motion/>} &  0.0 &  0.0 &  0.0 &  0.0 & 14.6 & 18.2 &  9.3 \\
    Qwen2.5-VL-7B & w/  \texttt{<motion/>} & \textbf{24.1} & \textbf{22.3} & \textbf{23.8} & \textbf{23.4} & \textbf{17.1} & \textbf{21.4} & \textbf{12.7} \\
\bottomrule
\end{tabular}
\end{table}
}

%% file: main.bib
@String(ICCV  = {Int. Conf. Comput. Vis.})

@String(NeurIPS = {Adv. Neural Inform. Process. Syst.})

@String(ICCV  = {ICCV})

@String(NeurIPS = {NeurIPS})

@article{meng2025open,
  title={Open-o3 Video: Grounded Video Reasoning with Explicit Spatio-Temporal Evidence},
  author={Meng, Jiahao and Li, Xiangtai and Wang, Haochen and Tan, Yue and Zhang, Tao and Kong, Lingdong and Tong, Yunhai and Wang, Anran and Teng, Zhiyang and Wang, Yujing and others},
  journal={arXiv preprint arXiv:2510.20579},
  year={2025}
}

@article{gu2025thinking,
  title={Thinking With Bounding Boxes: Enhancing Spatio-Temporal Video Grounding via Reinforcement Fine-Tuning},
  author={Gu, Xin and Zhang, Haoji and Fan, Qihang and Niu, Jingxuan and Zhang, Zhipeng and Zhang, Libo and Chen, Guang and Chen, Fan and Wen, Longyin and Zhu, Sijie},
  journal={arXiv preprint arXiv:2511.21375},
  year={2025}
}

@article{feng2025video,
  title={Video-R1: Reinforcing Video Reasoning in MLLMs},
  author={Feng, Kaituo and Gong, Kaixiong and Li, Bohao and Guo, Zonghao and Wang, Yibing and Peng, Tianshuo and Wu, Junfei and Zhang, Xiaoying and Wang, Benyou and Yue, Xiangyu},
  journal={arXiv preprint arXiv:2503.21776},
  year={2025}
}

@inproceedings{wang2025videorft,
  title={VideoRFT: Incentivizing Video Reasoning Capability in MLLMs via Reinforced Fine-Tuning},
  author={Wang, Qi and Yu, Yanrui and Yuan, Ye and Mao, Rui and Zhou, Tianfei},
  booktitle={The Thirty-ninth Annual Conference on Neural Information Processing Systems},
  year={2025}
}

@article{li2025videochat,
  title={VideoChat-R1: Enhancing Spatio-temporal Perception via Reinforcement Fine-tuning},
  author={Li, Xinhao and Yan, Ziang and Meng, Desen and Dong, Lu and Zeng, Xiangyu and He, Yinan and Wang, Yali and Qiao, Yu and Wang, Yi and Wang, Limin},
  journal={arXiv preprint arXiv:2504.06958},
  year={2025}
}

@inproceedings{wang2025videorts,
  title={Video-rts: Rethinking reinforcement learning and test-time scaling for efficient and enhanced video reasoning},
  author={Wang, Ziyang and Yoon, Jaehong and Yu, Shoubin and Islam, Md Mohaiminul and Bertasius, Gedas and Bansal, Mohit},
  booktitle={Proceedings of the 2025 Conference on Empirical Methods in Natural Language Processing},
  pages={28114--28128},
  year={2025}
}

@inproceedings{park2025deepvideo,
  title={DeepVideo-R1: Video Reinforcement Fine-Tuning via Difficulty-aware Regressive GRPO},
  author={Park, Jinyoung and Na, Jeehye and Kim, Jinyoung and Kim, Hyunwoo J},
  booktitle={NeurIPS},
  year={2025}
}

@inproceedings{wang2025time,
  title={Time-R1: Post-Training Large Vision Language Model for Temporal Video Grounding},
  author={Wang, Ye and Wang, Ziheng and Xu, Boshen and Du, Yang and Lin, Kejun and Xiao, Zihan and Yue, Zihao and Ju, Jianzhong and Zhang, Liang and Yang, Dingyi and others},
  booktitle={The Thirty-ninth Annual Conference on Neural Information Processing Systems},
  year={2025}
}

@article{ouyang2025spacer,
  title={SpaceR: Reinforcing MLLMs in Video Spatial Reasoning},
  author={Ouyang, Kun and Liu, Yuanxin and Wu, Haoning and Liu, Yi and Zhou, Hao and Zhou, Jie and Meng, Fandong and Sun, Xu},
  journal={arXiv preprint arXiv:2504.01805},
  year={2025}
}

@inproceedings{fan2025grit,
  title={GRIT: Teaching MLLMs to Think with Images},
  author={Fan, Yue and He, Xuehai and Yang, Diji and Zheng, Kaizhi and Kuo, Ching-Chen and Zheng, Yuting and Guan, Xinze and Wang, Xin Eric},
  booktitle={The Thirty-ninth Annual Conference on Neural Information Processing Systems},
  year={2025}
}

@article{wang2025vgr,
  title={VGR: Visual Grounded Reasoning},
  author={Wang, Jiacong and Kang, Zijian and Wang, Haochen and Jiang, Haiyong and Li, Jiawen and Wu, Bohong and Wang, Ya and Ran, Jiao and Liang, Xiao and Feng, Chao and others},
  journal={arXiv preprint arXiv:2506.11991},
  year={2025}
}

@article{zhang2025deepeyes,
  title={DeepEyes: Incentivizing Thinking with Images via Reinforcement Learning},
  author={Zheng, Ziwei and Yang, Michael and Hong, Jack and Zhao, Chenxiao and Xu, Guohai and Yang, Le and Shen, Chao and Yu, Xing},
  journal={arXiv preprint arXiv:2505.14362},
  year={2025}
}

@inproceedings{wang2025treevgr,
  title={Traceable Evidence Enhanced Visual Grounded Reasoning: Evaluation and Method},
  author={Wang, Haochen and Li, Xiangtai and Huang, Zilong and Wang, Anran and Wang, Jiacong and Zhang, Tao and Bai, Sule and Kang, Zijian and Feng, Jiashi and Zhuochen, Wang and others},
  booktitle={The Fourteenth International Conference on Learning Representations},
  year={2026}
}

@misc{openai2025o3,
  title={Introducing o3 and o4-mini},
  author={OpenAI},
  year={2025},
  howpublished={\url{https://openai.com/index/introducing-o3-and-o4-mini/}}
}

@article{bai2025qwen,
  title={Qwen2.5-VL Technical Report},
  author={Bai, Shuai and Chen, Keqin and Liu, Xuejing and Wang, Jialin and Ge, Wenbin and Song, Sibo and Dang, Kai and Wang, Peng and Wang, Shijie and Tang, Jun and others},
  journal={arXiv preprint arXiv:2502.13923},
  year={2025}
}

@article{zheng2025group,
  title={Group Sequence Policy Optimization},
  author={Zheng, Chujie and Liu, Shixuan and Li, Mingze and Chen, Xiong-Hui and Yu, Bowen and Gao, Chang and Dang, Kai and Liu, Yuqiong and Men, Rui and Yang, An and others},
  journal={arXiv preprint arXiv:2507.18071},
  year={2025}
}

@inproceedings{fu2024videomme,
  title={Video-mme: The first-ever comprehensive evaluation benchmark of multi-modal llms in video analysis},
  author={Fu, Chaoyou and Dai, Yuhan and Luo, Yongdong and Li, Lei and Ren, Shuhuai and Zhang, Renrui and Wang, Zihan and Zhou, Chenyu and Shen, Yunhang and Zhang, Mengdan and others},
  booktitle={Proceedings of the IEEE/CVF conference on computer vision and pattern recognition},
  pages={24108--24118},
  year={2025}
}

@article{hong2025worldsense,
  title={WorldSense: Evaluating Real-world Omnimodal Understanding for Multimodal LLMs},
  author={Hong, Jack and Yan, Shilin and Cai, Jiayin and Jiang, Xiaolong and Hu, Yao and Xie, Weidi},
  journal={arXiv preprint arXiv:2502.04326},
  year={2025}
}

@inproceedings{cho2025perceptionlm,
  title={PerceptionLM: Open-Access Data and Models for Detailed Visual Understanding},
  author={Cho, Jang Hyun and Madotto, Andrea and Mavroudi, Effrosyni and Afouras, Triantafyllos and Nagarajan, Tushar and Maaz, Muhammad and Song, Yale and Ma, Tengyu and Hu, Shuming and Jain, Suyog and others},
  booktitle={The Thirty-ninth Annual Conference on Neural Information Processing Systems},
  year={2025}
}

@inproceedings{wang2023vstar,
  title={VSTAR: A video-grounded dialogue dataset for situated semantic understanding with scene and topic transitions},
  author={Wang, Yuxuan and Zheng, Zilong and Zhao, Xueliang and Li, Jinpeng and Wang, Yueqian and Zhao, Dongyan},
  booktitle={Proceedings of the 61st Annual Meeting of the Association for Computational Linguistics (Volume 1: Long Papers)},
  pages={5036--5048},
  year={2023}
}

@inproceedings{trove,
title={{TR}oVe: Discovering Error-Inducing Static Feature Biases in Temporal Vision-Language Models},
author={Maya Varma and Jean-Benoit Delbrouck and Sophie Ostmeier and Akshay S Chaudhari and Curtis Langlotz},
booktitle={The Thirty-ninth Annual Conference on Neural Information Processing Systems},
year={2025},
}

@InProceedings{escalatorproblem,
    author    = {Zhang, Xiantao},
    title     = {The Escalator Problem: Identifying Implicit Motion Blindness in AI for Accessibility},
    booktitle = {Proceedings of the IEEE/CVF International Conference on Computer Vision (ICCV) Workshops},
    month     = {October},
    year      = {2025},
    pages     = {6635-6643}
}

@article{liu2024tempcompass,
  title={Tempcompass: Do video llms really understand videos?},
  author={Liu, Yuanxin and Li, Shicheng and Liu, Yi and Wang, Yuxiang and Ren, Shuhuai and Li, Lei and Chen, Sishuo and Sun, Xu and Hou, Lu},
  journal={arXiv preprint arXiv:2403.00476},
  year={2024}
}

@inproceedings{hong2025motionbench,
  title={Motionbench: Benchmarking and improving fine-grained video motion understanding for vision language models},
  author={Hong, Wenyi and Cheng, Yean and Yang, Zhuoyi and Wang, Weihan and Wang, Lefan and Gu, Xiaotao and Huang, Shiyu and Dong, Yuxiao and Tang, Jie},
  booktitle={Proceedings of the Computer Vision and Pattern Recognition Conference},
  pages={8450--8460},
  year={2025}
}

@inproceedings{li2024mvbench,
  title={Mvbench: A comprehensive multi-modal video understanding benchmark},
  author={Li, Kunchang and Wang, Yali and He, Yinan and Li, Yizhuo and Wang, Yi and Liu, Yi and Wang, Zun and Xu, Jilan and Chen, Guo and Luo, Ping and others},
  booktitle={Proceedings of the IEEE/CVF Conference on Computer Vision and Pattern Recognition},
  pages={22195--22206},
  year={2024}
}

@misc{vonwerra2022trl,
  author       = {Leandro von Werra and Younes Belkada and Lewis Tunstall and Edward Beeching and Tristan Thrush and Nathan Lambert and Shengyi Huang and Kashif Rasul and Quentin Gallou{\'e}dec},
  title        = {{TRL}: Transformer Reinforcement Learning},
  year         = {2020},
  publisher    = {GitHub},
  journal      = {GitHub repository},
  howpublished = {\url{https://github.com/huggingface/trl}}
}

@InProceedings{llarva2025,
  title = 	 {LLARVA: Vision-Action Instruction Tuning Enhances Robot Learning},
  author =       {Niu, Dantong and Sharma, Yuvan and Biamby, Giscard and Quenum, Jerome and Bai, Yutong and Shi, Baifeng and Darrell, Trevor and Herzig, Roei},
  booktitle = 	 {Proceedings of The 8th Conference on Robot Learning},
  pages = 	 {3333--3355},
  year = 	 {2025},
  editor = 	 {Agrawal, Pulkit and Kroemer, Oliver and Burgard, Wolfram},
  volume = 	 {270},
  series = 	 {Proceedings of Machine Learning Research},
  month = 	 {06--09 Nov},
  publisher =    {PMLR},
  url = 	 {https://proceedings.mlr.press/v270/niu25a.html},

}

@misc{videomolmo2025,
      title={VideoMolmo: Spatio-Temporal Grounding Meets Pointing},
      author={Ghazi Shazan Ahmad and Ahmed Heakl and Hanan Gani and Abdelrahman Shaker and Zhiqiang Shen and Ranjay Krishna and Fahad Shahbaz Khan and Salman Khan},
      year={2025},
      eprint={2506.05336},
      archivePrefix={arXiv},
      primaryClass={cs.CV},
url={https://arxiv.org/abs/2506.05336},
}

@article{bai2025qwen3,
  title={Qwen3-vl technical report},
  author={Bai, Shuai and Cai, Yuxuan and Chen, Ruizhe and Chen, Keqin and Chen, Xionghui and Cheng, Zesen and Deng, Lianghao and Ding, Wei and Gao, Chang and Ge, Chunjiang and others},
  journal={arXiv preprint arXiv:2511.21631},
  year={2025}
}
